\documentclass[onecolumn,3p]{elsarticle}

\usepackage{natbib}
\usepackage{amssymb}
\usepackage{times}
\usepackage{helvet}
\usepackage{courier}
\usepackage{verbatim}

\usepackage[pdftex]{color}
\usepackage{caption}
\usepackage{graphicx}

\usepackage{latexsym}
\usepackage{float}

\makeatletter
\newif\if@restonecol
\makeatother

\usepackage[ruled,vlined,linesnumbered]{algorithm2e}

\journal{Artificial Intelligence}

\usepackage{multirow}
\usepackage{array}
\newcommand{\PreserveBackslash}[1]{\let\temp=\\#1\let\\=\temp}
\newcolumntype{C}[1]{>{\PreserveBackslash\centering}p{#1}}
\newcolumntype{R}[1]{>{\PreserveBackslash\raggedleft}p{#1}}
\newcolumntype{L}[1]{>{\PreserveBackslash\raggedright}p{#1}}

\begin{document}
\sloppy

\begin{frontmatter}

\title{Clause Vivification by Unit Propagation in CDCL SAT Solvers\tnoteref{t1}}


\author[hust,upjv]{Chu-Min Li}\ead{chu-min.li@u-picardie.fr}
\author[hust]{Fan Xiao}\ead{fanxiao@hust.edu.cn}
\author[hust]{Mao Luo}\ead{maoluo@hust.edu.cn}
\author[iiia]{Felip Many\`a}\ead{felip@iiia.csic.es}
\author[hust]{Zhipeng L{\"u}}\ead{zhipeng.lv@hust.edu.cn}
\author[upjv]{Yu Li}\ead{yu.li@u-picardie.fr}

\address[hust]{School of Computer Science, Huazhong University of Science and Technology, Wuhan, China.}
\address[upjv]{MIS, University of Picardie Jules Verne, Amiens, France.}
\address[iiia]{Artificial Intelligence Research Institute (IIIA), Spanish Scientific Research Council (CSIC), Bellaterra, Spain.}

\begin{abstract}

Original and learnt clauses in Conflict-Driven Clause Learning (CDCL) SAT solvers often contain redundant literals. This may have a negative impact on performance because redundant literals may deteriorate both the effectiveness of Boolean constraint propagation and the quality of subsequent learnt clauses. To overcome this drawback, we propose a clause vivification approach that eliminates redundant literals by applying unit propagation. The proposed clause vivification is activated before the SAT solver triggers some selected restarts, and only affects a subset of original and learnt clauses, which are considered to be more relevant according to metrics like the literal block distance (LBD). Moreover, we conducted an empirical investigation with instances coming from the hard combinatorial and application categories of recent SAT competitions. The results show that a remarkable number of additional instances are solved when the proposed approach is incorporated into five of the best performing CDCL SAT solvers (Glucose, TC\underline{\hspace{2mm}}Glucose, COMiniSatPS, MapleCOMSPS and MapleCOMSPS\underline{\hspace{2mm}}LRB). More importantly, the empirical investigation includes an in-depth analysis of the effectiveness of clause vivification. It is worth mentioning that one of the SAT solvers described here was ranked first in the main track of SAT Competition 2017 thanks to the incorporation of the proposed clause vivification. 
That solver was further improved in this paper and won the bronze medal in the main track of SAT Competition 2018.
\end{abstract}


\end{frontmatter}

\section{Introduction}

In propositional logic, a variable $x$ may take the truth value 0 (false) or 1 (true). A literal $l$ is a variable
$x$ or its negation $\neg x$, a clause is  a disjunction of  literals, a CNF formula $\phi$ is a conjunction of clauses, and the size of a clause is the number of literals in it.
An assignment of truth values to the propositional variables satisfies a literal $x$ if $x$ takes the value~1
and satisfies a literal $\neg x$ if $x$ takes the value~0, satisfies a clause if it satisfies at least one of its literals, and satisfies a CNF formula if it satisfies all of its clauses. The empty clause, denoted by $\Box$, contains no literals and is unsatisfiable; it represents a conflict. A unit clause contains exactly one literal and is satisfied by assigning the appropriate truth value to the variable. 
 An assignment for a CNF formula $\phi$ is complete if each variable in $\phi$ has been assigned a value; otherwise, it is considered partial. 
 The SAT problem for a CNF formula $\phi$ is to find an assignment to the variables that satisfies all clauses of $\phi$. One of the interests of SAT is that many combinatorial problems can easily be encoded as a SAT problem. For example, Cook showed that any NP problem can be encoded to SAT in polynomial time \cite{cook1971complexity}, founding the theory of NP-completeness.


Because of the high expressive power of SAT and the progress made on SAT solving techniques, modern Conflict-Driven Clause Learning (CDCL) SAT solvers are routinely used as core
solving engines in many real-world applications. Their ability to solve challenging problems comes from the combination of different components: variable selection heuristics, unit clause propagation, clause learning, restarts, clause database management, data structures, pre- and inprocessing.

Formula simplification techniques applied during preprocessing have proven useful in enabling efficient SAT solving for real-world application domains (e.g.~\cite{BW03,EB05,PHS08}). The most successful preprocessing techniques include variants of bounded variable elimination, addition or elimination of redundant clauses, detection of subsumed clauses and suitable combinations of them. They aim mostly at reducing the number of clauses, literals and variables in the input formula.

More recently, interleaving formula simplification techniques with CDCL search has provided significant performance improvements. Among such {\em inprocessing techniques}~\cite{JHB12}, we mention local and recursive clause minimization~\cite{BKS04,SB09}, which remove redundant literals from learnt clauses immediately after their creation; clause vivification in a concurrent context~\cite{wieringa2013concurrent,van2012satuzk}; and on-the-fly clause subsumption \cite{han2009fly,hamadi2009learning}, which efficiently removes clauses subsumed by the resolvents derived during clause learning. 

In this paper, we focus on clause vivification, which consists in eliminating redundant literals from clauses. 
A clause is a logical consequence of a CNF $\phi$ if every solution of $\phi$ satisfies it.
Let $C=l_1 \vee l_2 \vee \cdots \vee l_k$ be a clause of $\phi$, $l_i$ be a literal of $C$ and $C\setminus l_i$ be the clause obtained from $C$ after removing $l_i$.  
If $C\setminus l_i$ is a logical consequence of $\phi$, then $l_i$ is said to be {\em redundant} in $C$ and should be eliminated from $C$. Indeed, from a problem solving pespective, $C\setminus l_i$ is always better than $C$ when $l_i$ is redundant in $C$. Nevertheless, identifying a redundant literal in a clause $C$ of $\phi$  is NP-hard in general. So, in this paper, we restrict clause vivification to the elimination of redundant literals that can be identified by applying unit clause propagation (or simply unit propagation), as described below. 

Solving a SAT problem $\phi$ amounts to satisfy every clause of $\phi$. We pay special attention to the unit clauses in $\phi$ (if any). 
Satisfying a unit clause $l$ implies to falsify the literal $\neg l$, which should be removed from each clause $C'$ containing it because $C'$ cannot satisfied by $\neg l$. If $C'$ becomes unit ($C'=l'$), then $l'$ is satisfied and $\neg l'$ is removed from the remaining clauses, and so on. 
This process is called unit propagation and is denoted by UP($\phi$), and continues until there is no unit clause in $\phi$ or a clause becomes empty. In the latter case, $\phi$ is proved to be unsatisfiable. 
For any literal $l_i$ in $C$,  if UP($\phi \cup \{\neg l_1, \ldots, \neg l_{i-1}, \neg l_{i+1}, \ldots, \neg l_k\})$ results in the empty clause, then $C\setminus l_i$ is a logical consequence of $\phi$ and $l_i$ is a redundant literal. Thus, $C$ should replaced by $C\setminus l_i$ in $\phi$. 

Clause vivification was independently proposed in~\cite{HS07} and~\cite{PHS08} to preprocess the input formula in SAT solvers\footnote{It is called {\em distillation} in \cite{HS07}.}. 
Unfortunately, it is not easy to make clause vivification effective because of the cost of unit propagation. This can be illustrated by the evolution of Lingeling, a frequently awarded solver in SAT competitions\footnote{www.satcompetition.org}: Clause vivification was implemented in Lingeling 271 \cite{biere2010lingeling}  in 2010 but was removed in 2012  because it did not have any observable impact on the runtime of the tested benchmarks~\cite{biere2012lingeling}. In the descriptions of Lingeling in subsequent  SAT competitions~\cite{biere2013lingeling,biere2014yet,biere2015lingeling,biere2016splatz}, clause vivification is not even mentioned.

Although clause vivification was proposed as a preprocessing technique in~\cite{PHS08}, the authors mentioned inprocessing clause vivification as future work. Actually, making inprocessing clause vivification effective is much harder. We quote a statement from three leading experts in the field to partly appreciate this hardness \cite{JHB12}:
\begin{quotation}
``However, developing and implementing sound inprocessing solvers in the presence of a wide range of different simplification techniques is highly non-trivial. It requires in-depth understanding on how different techniques can be combined together and interleaved with the CDCL algorithm in a satisfiability-preserving way."
\end{quotation}

A question that should be addressed in inprocessing clause vivification is to determine whether or not to apply vivification to learnt clauses, because each learnt clauses is a logical consequence of the input formula and can be removed when the clause database is reduced. To the best of our knowledge, none of the awarded CDCL SAT solvers in SAT Competition 2016 used clause vivification to eliminate redundant literals in learnt clauses by applying unit propagation. In particular, the solver Riss6, the silver medal winner of the main track, disabled the learnt clause vivification used in Riss 5.05, because it turned out to be ineffective for formulas of more recent years~\cite{MSW16}.

The main purpose of this paper is to design an effective and efficient approach to vivifying both original clauses and learnt clauses in pre- and inprocessing. Because of the difficulty stated above, we designed the proposed approach in an incremental way. 

Firstly, we limited vivification to inprocessing and only vivified a subset of learnt clauses selected using a heuristic
at some carefully defined moments during the search process. Moreover, each learnt clause was vivified at most once. This preliminary approach was presented in~\cite{LLXML17} and implemented in five of the best performing CDCL SAT solvers: Glucose~\cite{AS09}, COMiniSatPS~\cite{Oh16}, MapleCOMSPS~\cite{LOGCP16}, MapleCOMSPS\underline{\hspace{2mm}}LRB~\cite{LGPC16} and TC\underline{\hspace{2mm}}Glucose~\cite{moon2016chbr}. The experimental results show that the proposed vivification allows these solvers to solve a remarkable number of additional instances coming from the hard combinatorial and application categories of the SAT Competition 2014 and 2016. More importantly, we submitted four solvers based on our approach to SAT competition 2017 -- Maple\_LCM and three variants with some other techniques --  and won the gold medal of the main track. The four solvers solved 20, 18, 16 and 10 instances more than MapleCOMSPS\_LRB\_VSIDS\_2, the first solver without our approach, in the main track consisting of 350 industrial and hard combinatorial instances. In addition, the best two solvers in the main track of SAT competition 2018, Maple\_LCM\_Dist\_ChronoBT~\cite{Nadel2018} and Maple\_LCM\_Scavel~\cite{yang2018}, are developed from Maple\_LCM by integrating some other techniques. Furthermore, 
the best 13 solvers of the main track of SAT Competition 2018, developed by different author sets, all use our learnt clause vivification approach~\cite{sc2018}.

Secondly, in this paper, we extend the above preliminary approach after conducting a more extensive investigation that includes an in-depth analysis of the most important steps of the vivification process. In fact, our investigation indicates that the most crucial point is to determine which clauses must be vivified rather than to decide when vivification should be activated. So, we define a new clause selection heuristic that extends the vivification approach to original clauses and allows some clauses to be vivified more than once. Implemented in Maple\_LCM, the extended approach allows it to solve 34 more instances among the 1450 instances in the main track of the SAT Competition 2014, 2016 and 2017 (of which 20 more instances among the 350 instances in the main track of the SAT Competition 2017). The new solver, called Maple\_CV+ in this paper, participated in SAT competition 2018 under the name Maple\_CM~\cite{luo2018} and won the bronze medal of the main track.

Note that the benchmarks used in SAT competition were not known before the solvers were submitted.

The main contribution of this paper is that it definitely shows the benefits of vivifying original and learnt clauses during the search process. Furthermore, the empirical results indicate that the proposed approach is robust and is not necessary to carefully adjust parameters to obtain significant gains, even with an implementation that is not fully optimized. We quote a statement of two leading SAT solver developers \cite{audemard2012refining} to better appreciate the significance of our contributions:\\

\begin{quotation}
``We must also say, as a preliminary, that improving SAT solvers is often a cruel world. To give an idea, improving a solver by solving at least ten more instances (on a fixed set of benchmarks of a competition) is generally showing a critical new feature. In general, the winner of a competition is decided based on a couple of additional solved benchmarks."\\
\end{quotation}

The above statement was also quoted in \cite{LGPC16aaai} to appreciate the advances brought about by the CHB branching heuristic.

The paper is organized as follows: Section~\ref{preliminaries} gives some basic concepts about propositional satisfiability and CDCL SAT solvers.
Section~\ref{relatedWork} presents some related works on  elimination of redundant literals.
Section~\ref{mainSection} describes our learnt clause vivification approach, as well as how it is implemented in different CDCL SAT solvers.
Section~\ref{experiments} reports on the in-depth empirical investigation of the proposed clause vivification approach. Section~\ref{approachImprovement} extends clause vivification to original clauses during the search process, defines the conditions to re-vivify learnt and original clauses, and assesses the performance of these extensions.
Section~\ref{conclusions} contains the concluding remarks.

\section{Preliminaries} \label{preliminaries}

Consider two clauses $C'$ and $C$. If every literal of $C'$ is also a literal of $C$, then $C'$ is a sub-clause of $C$. A clause $C'$ subsumes a clause $C$ iff $C'$ is a sub-clause of $C$, because when $C'$ is satisfied, $C$ is also necessarily satisfied. For any literal $l$, it holds that $l \equiv \neg \neg l$.

A CNF formula is also represented as a set of clauses. A clause $C$ is redundant in a CNF formula $\phi$ if $C$ is a logical consequence of $\phi \setminus \{C\}$. Two CNF formulas $\phi$ and $\phi'$ are equivalent if they are satisfied by the same assignments. If $C'$ is a logical consequence of $\phi$ and $C'$ is a sub-clause of a clause $C$ of $\phi$, then replacing $C$ by $C'$ in $\phi$ results in a formula equivalent to $\phi$ that is easier to solve, because a solver considers fewer possibilities to satisfy $C'$.


Given a CNF formula $\phi$ and a literal $l$, we define $\phi\vert_l$ to be the CNF formula resulting from $\phi$ after removing the clauses containing an occurrence of $l$ and all the occurrences of $\neg l$. Unit Propagation (UP) can be recursively defined using the following two rules: (R1)~$UP(\phi)=\phi$ if $\phi$ does not contain any unit clause; and (R2)~$UP(\phi)=UP(\phi\vert_l)$ if there is a unit clause $l$ in $\phi$. 
 Literal $l$ is asserted by $UP(\phi)$ if
 $UP(\phi\vert_l)$ is computed.
 
In the sequel, $UP(\phi)$ denotes the CNF formula obtained from $\phi$ after repeatedly applying R2 until there is no unit clause or the empty clause is derived. Concretely, $UP(\phi)=\Box$ if the empty clause is derived. Otherwise, $UP(\phi)$ denotes the CNF formula in which all unit clauses have been propagated using R2. 
When $UP(\phi \cup \{l\})$ is computed, we say that  $l$ is propagated in $\phi$. 

The key idea behind vivifying a clause $C$ in a CNF formula $\phi$ is to identify a sub-clause $C'=l'_1 \vee \l'_2 \vee \cdots \vee l'_i$ of~$C$ such that UP$(\phi \cup \{\neg l'_1, \neg l'_2, \ldots, \neg l'_i\})=\Box$. In this case, $C'$ is a logical consequence of $\phi$, because UP$(\phi \cup \{\neg l'_1, \neg l'_2, \ldots, \neg l'_i\})=\Box$ means that $C'$ cannot be falsified when $\phi$ is satisfied. Hence, we can replace $C$ by $C'$ in $\phi$. In this way, we eliminate redundant literals from $C$ and obtain an easier formula to solve.

 A CDCL SAT solver~\cite{MarquesSakallah99,MMZZM01} performs a non-chronological backtrack search in the space of partial truth assignments. Concretely, the solver repeatedly picks a decision literal $l_i$ (for $i=1, 2, \ldots$) and applies unit propagation in $\phi\cup \{l_1, l_2, \ldots, l_i\}$ (i.e., it computes $UP(\phi\cup \{l_1, l_2, \ldots, l_i\})$) until the empty formula or the empty clause are derived. If the empty clause is derived, the reasons of the conflict are analyzed and a clause (nogood) is learnt using a particular method, usually the First UIP (Unique Implication Point) scheme~\cite{ZMMM01}. The learnt clause is then added to the clause database. Since too many learnt clauses slow down the solver and may overflow the available memory, the solver periodically removes a subset of learnt clauses using a particular clause deletion strategy.  

Let $\phi$ be a CNF formula, let $l_i$ be the $i^{th}$ decision literal picked by a CDCL SAT solver, and let $\phi_{i}=UP(\phi \cup \{l_1, l_2, \ldots, l_{i}\})$. We say that $l_i$ and all the literals asserted when computing $UP(\phi_{i-1} \cup \{l_i\})$ belong to level $i$. Literals asserted in $UP(\phi)$ do not depend on any decision literal and form level 0.
When UP derives the empty clause and a learnt clause is extracted from the conflict analysis, the solver cancels the literal assertions in the reverse order until the level where the learnt clause contains one non-asserted literal, and continues the search from that level after propagating the non-asserted literal.
Under certain conditions, the solver cancels the literal assertions until level 0 and restarts the search from level 0. Algorithm \ref{CDCL} in Section \ref{mainSection} depicts a generic CDCL SAT solver.

The literals of a learnt clause $C$ are partitioned w.r.t.~their assertion level. The number of sets in the partition is called
the Literal Block Distance (LBD) of $C$~\cite{AS09}. As shown in~\cite{AS09}, LBD measures the quality of learnt clauses. Clauses with small LBD values are considered to be more relevant. The best performing CDCL SAT solvers of the last SAT competitions use LBD as a measure of the quality of learnt clauses to determine which clauses must be removed or retained. Moreover, solvers like Glucose and its descendants use the LBD of recent learnt clauses to decide when a restart must be triggered.

\begin{figure}[!h]
	\centering
	\includegraphics[width=0.85\textwidth]{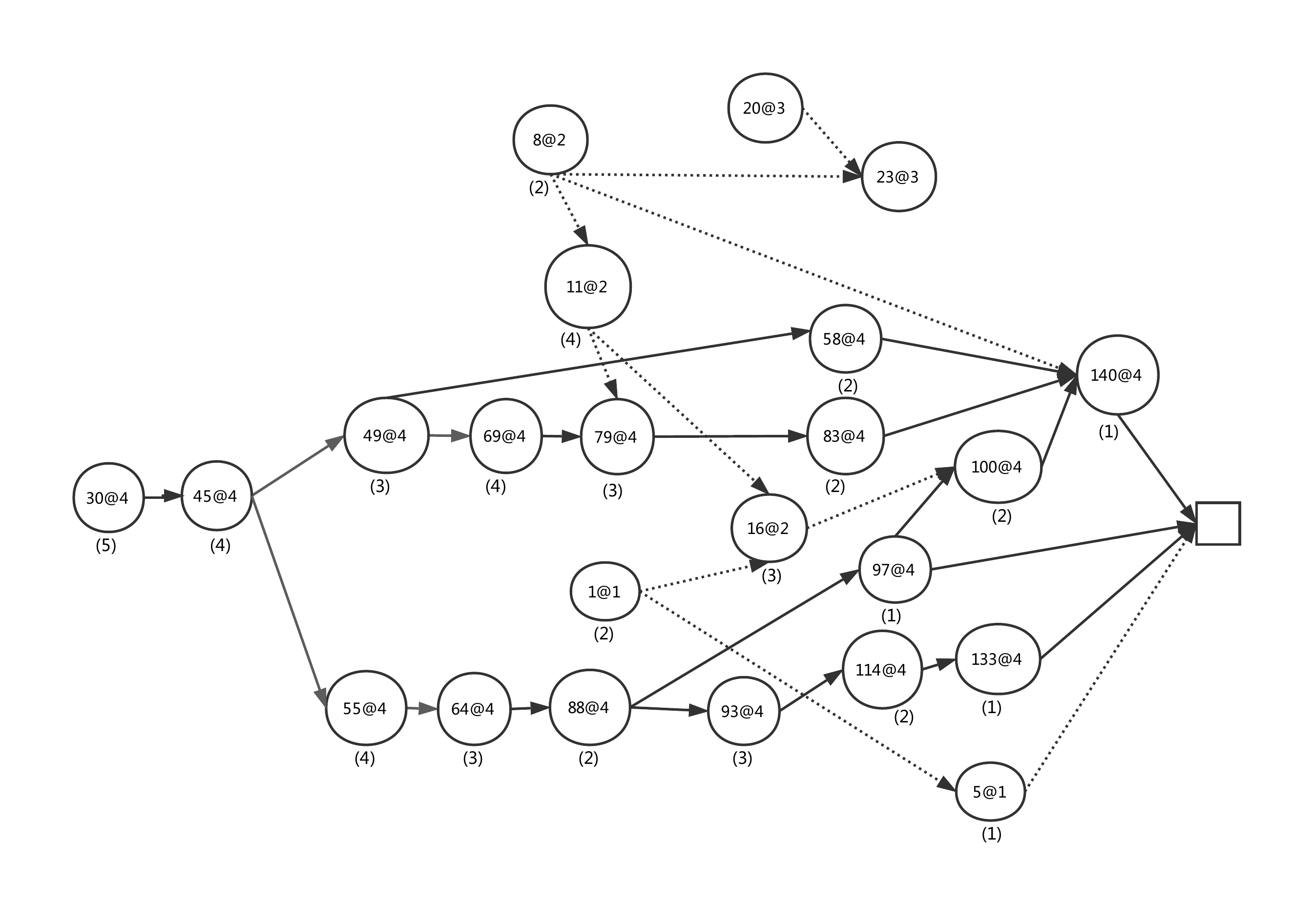}
	\caption{\small A complete implication graph. The number in parentheses below a vertex represents the distance from the vertex to $\Box$.  
	The arrows involving at least one vertex of a lower level are dotted.
	}
	\label{implicationFig}
\end{figure}

A CDCL SAT solver essentially constructs a (directed acyclic) implication graph $G$ as follows. Given a CNF~$\phi$ without unit clauses, a decision literal is a vertex of $G$. If there is a clause 
$\neg l_1\vee \neg l_2\vee \cdots \vee \neg l_{k-1} \vee l_k$ 
that becomes unit because $l_1, l_2, \ldots, l_{k-1}$ are satisfied (i.e., $ l_1, l_2, \ldots, l_{k-1}$ are in $G$), then vertex $l_k$ and arrows $(l_1, l_k), (l_2, l_k), \ldots, (l_{k-1}, l_k)$ are added to the graph. Figure~\ref{implicationFig} shows an example of implication graph. 
Following the terminology of~\cite{silva2003grasp}, the vertex name $a@b$, where $a$ and $b$ are positive integers, means that literal $l_a$ is asserted at decision level $b$. A variable $x_i$ is associated with two literals: $\neg x_i=l_{2i-1}$ and $x_i = l_{2i}$. Thus, an even number $a$ in the graph represents a positive literal and an odd number a negative literal. Therefore, $l_a=x_{\frac{a}{2}}$ if $a$ is even and $l_a=\neg x_{\frac{a+1}{2}}$ if $a$ is odd. 

A vertex in an implication graph represents a satisfied literal $l$ and also identifies a clause which is the reason of the satisfaction of $l$. A vertex without any predecessor identifies a unit clause (i.e., a decision literal). For example, vertex $30@4$ in Figure~\ref{implicationFig} identifies the unit clause $l_{30}$.  A vertex with incoming arrows identifies a non-unit clause. For example, vertex $140@4$ identifies, together with the incoming arrows, clause $\neg l_8 \vee \neg l_{58} \vee  \neg l_{83} \vee \neg l_{100} \vee l_{140}$, which is the reason to satisfy $l_{140}$. Note that the predecessors are negated in the clause, because $\neg l_8 \vee \neg l_{58} \vee  \neg l_{83} \vee \neg l_{100} \vee l_{140} \equiv l_8 \wedge l_{58} \wedge  l_{83} \wedge  l_{100} \rightarrow l_{140}$.

A clause vivification procedure executing UP$(\phi \cup \{\neg l'_1, \neg l'_2, \ldots, \neg l'_i\})$ also constructs an implication graph, where literals in $\{\neg l'_1, \neg l'_2, \ldots, \neg l'_i\}$ are considered as successive decision literals.

The length of a path between two vertices in an implication graph is the number of arrows in the path, and the distance between two vertices is the length of the shortest path between the two vertices.
A UIP in an implication graph deriving a conflict is a vertex through which all paths from the last decision literal to the conflict go. For example, the implication graph in Figure \ref{implicationFig} has four levels, the last decision literal is $l_{30}$, and there are two UIPs: $l_{30}$ and $l_{45}$, of which the first UIP (counting from the conflict) is $l_{45}$. Using the first UIP scheme, the learnt clause $\neg l_{45} \vee \neg l_8 \vee \neg l_5 \vee \neg l_{16} \vee \neg l_{11}$ is derived from the implication graph, consisting of the negation of all literals of a lower level whose distance to a literal of the highest level is 1, together with the negation of the first UIP. The learnt clause is recorded to avoid the re-construction of the implication graph when the literals in the learnt clause are falsified, because the learnt clause is already falsified in this case.

A clause plays a role in a search process only when it becomes unit. This fact is exploited in the {\em two-literal watching} technique \cite{MMZZM01} to speed up UP. Using this technique, a solver watches only two literals of each clause $C$, and does nothing on $C$ when other literals of $C$ are satisfied or falsified, because when $C$ becomes unit, one of the two watched  literals is necessarily falsified. Concretely, when one of the two watched literals is falsified, the technique inspects $C$ to see if it becomes unit or falsified. If not, the solver replaces the falsified literal by another non-falsified literal and watches it instead. In this way, backtracking just needs to cancel the variable assignments.

\section{Related Work on Elimination of Redundant Literals} \label{relatedWork}

Eliminating redundant literals in clauses (see e.g.~\cite{EB05,HS07,PHS08,SB09,wotzlaw2013effectiveness,marques2000algebraic,han2009fly,hamadi2009learning,wieringa2013concurrent}) before and during the search is crucial for the performance of CDCL SAT solvers for several reasons: (i)~shorter clauses need less memory;  (ii)~shorter clauses are easier to become unit, and thus increase the power of unit propagation; and (iii)~shorter clauses can lead to shorter learnt clauses.

The most effective approach to remove redundant literals in learnt clauses probably is  {\em recursive clause minimization}~\cite{SB09}. Let $C=l_1 \vee l_2 \vee \cdots \vee l_k$ be a newly created learnt clause. For each literal $l_i$ of $C$, recursive clause minimization checks, in the implication graph allowing to derive $C$, if every path arriving at $\neg l_i$ contains the negation of a literal of $C$. If so, $l_i$ is redundant and is removed from $C$.
 MiniSAT~1.13 implements recursive clause minimization and applies it to each new learnt clause. This approach is now part of many CDCL SAT solvers and frequently removes more than 30\%  of literals on average~\cite{SB09}. For example, the implication graph in Figure \ref{implicationFig}  allows to derive the learnt clause $\neg l_{45} \vee \neg l_8 \vee \neg l_5 \vee \neg l_{16} \vee \neg l_{11}$. Recursive clause minimization eliminates $\neg l_{11}$ from the clause because the clause contains $\neg l_8$, while $l_8$ is in the unique path arriving at $l_{11}$.

Another effective inprocessing clause minimization approach is based on binary implication graphs involving only clauses 
containing two literals~\cite{heule2011efficient}. Given a CNF formula $\phi$, the algorithm first performs a depth-first search over the binary implication graph to assign a time stamp to each literal in the graph. It then uses these time stamps to discover and remove various kinds of redundancy, including redundant literals in learnt clauses. The algorithm is linear in the total number of literals in $\phi$ and is realized in a function named $Stamp$ in Lingeling \cite{biere2010lingeling} and MapleCOMSPS\underline{\hspace{2mm}}LRB. This function is executed upon some restarts when it is used as an inprocessing clause minimization technique.

Solvers such as Glucose and MapleCOMSPS apply a restriction of the resolution rule to shorten a newly created learnt clause $C=l_1 \vee l_2 \vee \cdots \vee l_k$ whose LBD or size  is smaller than a prefixed limit. The first literal $l_1$ of $C$ is called the asserting literal. For any $i>1$, if $l_1 \vee \neg l_i$ is a clause in $\phi$, then Glucose and MapleCOMSPS remove $l_i$ from $C$. 

Conflict analysis consists in performing successive resolution steps from a conflicting clause. At each step, a new resolvent is derived from two clauses $C'$ and $C''$, where $C'$ is the conflicting clause or a previous resolvent and $C''$ is an existing clause. If $C''=l \vee C$ and $C'=\neg l \vee D$, where $l$ is a literal, and $C$ and $D$ are disjunction of literals with $D \subseteq C$,  the resolvent of $C'$ and $C''$ is $C$ and subsumes $C''$, meaning that $l$ is redundant in $C''$ and can be removed from $C''$. This approach allows to minimize both original and learnt clauses during the search and was proposed independently in \cite{han2009fly} and \cite{hamadi2009learning}. It was implemented in CryptoMinisat 2.5.0 \cite{soos2010cryptominisat} for on-the-fly clause improvement.


In~\cite{PHS08}, the following rules are applied in a vivification procedure after sorting the literals of a clause $C=l_1 \vee l_2 \vee \cdots \vee l_k$ of the input formula $\phi$ using a  MOMS-style \cite{jeroslow1990solving} ordering:

\begin{enumerate}
\item If for some $i \in \{1, 2, \ldots, k-1\}$ and some $j>i$, $UP((\phi\setminus \{C\}) \cup \{\neg l_1, \neg l_2, \ldots, \neg l_i\})$ deduces $l_j$, then $\phi \leftarrow (\phi\setminus \{C\}) \cup \{l_1 \vee l_2 \vee \ldots \vee l_i \vee l_j\}$.
\item If for some $i \in \{1, 2, \ldots, k-1\}$ and some $j>i$, $UP((\phi\setminus \{C\}) \cup \{\neg l_1, \neg l_2, \ldots, \neg l_i\})$ deduces $\neg l_j$, then $\phi \leftarrow (\phi\setminus \{C\}) \cup \{l_1 \vee l_2 \vee \ldots \vee l_i \vee \cdots \vee l_{j-1} \vee l_{j+1} \vee \cdots \vee l_k\}$.
\item If for some $i \in \{1, 2, \ldots, k-1\}$, $UP((\phi\setminus \{C\}) \cup \{\neg l_1, \neg l_2, \ldots, \neg l_i\})=\Box$, then extract a nogood $C_l$ from the conflict using the First UIP scheme. If $C_l$  subsumes  $l_1 \vee l_2 \vee \ldots \vee l_i$, then $\phi \leftarrow (\phi\setminus \{C\}) \cup C_l$;  otherwise, $\phi \leftarrow (\phi\setminus \{C\}) \cup \{l_1 \vee l_2 \vee \ldots \vee l_i\}$. Additionally, when $|C_l| < |C|$, $C_l$ is also added to $\phi$.

\end{enumerate}

Independently, a similar approach called distillation was proposed in~\cite{HS07}.

Note that the previous clause vivification or distillation is quite different from recursive clause minimization of~\cite{SB09} 
and binary implication graph based clause minimization of~\cite{heule2011efficient}, despite that all eliminate redundant literals from clauses.
In fact, recursive clause minimization deals with an existing implication graph where the negation of each literal of the clause to be minimized is already propagated. In the binary implication graph based clause minimization, the same binary implication graph is constructed to minimize all the clauses. So, the cost of both approaches is relatively small. However, in clause vivification or distillation, a different implication graph is constructed, by propagating the negation of the literals of the clause, for each clause to be minimized. Hence, the cost of clause vivification or distillation is quite high. For this reason, clause vivification or  distillation in~\cite{PHS08,HS07} is only proposed to preprocess the input CNF formula before starting the search.

Different inprocessing clause vivification approaches have been proposed in the literature~\cite{van2012satuzk,wotzlaw2013effectiveness,wieringa2013concurrent,Biere2017}.  In~\cite{van2012satuzk,wotzlaw2013effectiveness} only the most active clauses are vivified and only 10\% of runtime is used for inprocessing. The approach in~\cite{wieringa2013concurrent} is used in 
the context of parallel solvers, where a concurrent thread was used to repeatedly select and vivify the best learnt clause (according to some criteria such as clause size or LBD) among the newest 1000 learnt clauses. In~\cite{Biere2017}, which presents the SAT solver Cadical (version 2017), only  irredundant clauses are vivified. In summary, all these approaches vivify a quite small set of clauses, presumably because of the high cost of clause vivification. Consequently, their benefits are not so compelling. 

After the publication of the IJCAI version of this paper~\cite{LLXML17}, our learnt clause vivification approach was implemented in the version of the SAT solver Glucose~\cite{glucose2018} submitted to SAT Competition 2018, while previous versions of Glucose did not use any clause vivification. Another solver submitted to SAT Competition 2018, Cadical-2018~\cite{Biere2018}, also implements learnt clause vivification in the spirit of the ideas presented in~\cite{LLXML17}.
Note that Glucose was a frequently awarded solver in SAT competitions and recent SAT competitions have a Glucose-hack track. Cadical might be considered as an improved version of the solver Lingeling by the same author.  Lingeling was also a frequently awarded solver in SAT competitions.

The purpose of this paper is to show that clause vivification applied to both original and learnt clauses in a solver can be highly effective when its application is guided by some clever principles.

\section{The Proposed Clause Vivification Approach \label{mainSection}}

We first describe the general principle of the proposed approach, and then present its implementation in the solvers Glucose, TC\underline{\hspace{2mm}}Glucose, COMiniSatPS, MapleCOMSPS and MapleCOMSPS\underline{\hspace{2mm}}LRB.

\subsection{General Principles} \label{original}

In order to make inprocessing clause vivification effective and efficient, we address the following questions:

\begin{enumerate}

\item When should we activate clause vivification?
Clause vivification should be activated at level 0 to ensure that the vivification is independent of any branching decision. In other words, it should be activated upon a restart. 
Hence, the relevant questions are: should we activate clause vivification in every restart? If not, how do we determine the restarts upon which we activate clause vivification?

\item Should we vivify each clause? If not, which clauses should be vivified?

\item Clause vivification depends on the ordering selected to propagate the  negation of the literals of a clause; different orderings may derive vivified clauses of different lengths. What is the best order to propagate the literals of a clause when we apply vivification?
\end{enumerate}

The most satisfactory answers to these questions may depend on other techniques implemented in the solver.  Nevertheless, we want to argue the following general principles for guiding the implementation of inprocessing clause vivification:
\begin{enumerate}

\item It is not necessary to activate clause vivification at each restart. In fact, the number of new learnt clauses per restart may not be sufficient to easily deduce a conflict by unit propagation.  Roughly, we can say that it depends on the number of clauses which were learnt since the last learnt clause vivification ($nbNewLearnts$), and the number of clause vivifications performed so far ($\sigma$). We define later function $liveRestart(nbNewLearnts, \sigma)$ to determine whether or not clause vivification has to be activated at the beginning of a restart.
%

\item  Chanseok Oh (\cite{Oh16})   demonstrated empirically that learnt clauses with high LBD values are not very useful to solve practical SAT instances. Indeed, the LBD of a learnt clause is correlated to the number of decisions needed to falsify the clause, meaning that it is much harder to reduce the LBD of a learnt clause than to reduce its size. Moreover, clauses with high LBD values are generally long and need more unit propagations to be vivified. In Section~\ref{experiments}, we will provide empirical evidence that vivifying clauses with high LBD values is very costly and useless. Therefore, we propose to use LBD values as a measure to determine whether or not a clause should be vivified and the proposal is to vivify clauses with small LBD values. 
In practice, we define the function $liveClause(C)$ to determine whether or not the clause $C$ has to be vivified.

\item We should propagate the negation of the literals of a clause $C$ in their current ordering to vivify the clause. In the next subsection, we will explain why this is probably the best ordering, which is also confirmed by the reported experimental investigation. Anyway, we assume the existence of function $sort(C)$ for convenience, which sorts the literals of clause $C$ before applying clause vivification to $C$.

\end{enumerate}

The above general principle can be summarized by saying that clause vivification should be applied to selected relevant clauses when a restart is triggered if a considerable number of conflicts have been detected since the last clause vivification. However, we will provide evidence that the most crucial aspect is to determine which clauses should be vivified or re-vivified rather than when they should be vivified or re-vivified.

Algorithm~\ref{CDCL} is a generic CDCL SAT solver. It calls  function vivifyIfPromising$(\phi, nbNewLearnts, \sigma)$ (i.e., Algorithm~\ref{minimization}) at the beginning of each restart to apply clause vivification if function liveRestart$(nbNewLearnts, \sigma)$ returns true.  Function liveRestart$(nbNewLearnts, \sigma)$ will be defined when Algorithm~\ref{minimization} is implemented in a real CDCL SAT solver. Function vivify$(\phi)$, which realizes clause vivification in $\phi$, is defined in Algorithm~\ref{vivifying}. 

In Algorithm~\ref{vivifying},  function liveClause$(C)$ will be defined when Algorithm~\ref{vivifying} is implemented in a real CDCL SAT solver, and function sort$(C)$ will be defined in the next subsection. 

\begin{algorithm}[!h]
\KwIn{$\phi$: A CNF formula with original and learnt clauses.}
\KwOut{SATISFIABLE or UNSATISFIABLE}
\Begin{
	$\phi \leftarrow$ preprocessing$(\phi)$:\\
	\While {true} {
		$currentLevel \leftarrow 0$;  /* start or restart search */\\
		$\phi \leftarrow$ vivifyIfPromising$(\phi, nbNewLearnts, \sigma)$; /* call Algorithm~\ref{minimization} */\\
		\While {true} {
			$cl \leftarrow$ UP$(\phi)$; /* all variables assigned by UP are recorded with $currentLevel$ */\\
			\If {$cl$ is a falsified clause}{
				\If {$currentLevel == 0$}{
					return UNSATISFIABLE;
				}
				\Else{
					$newLearntClause \leftarrow$ analyze$(cl)$; /* conflict analysis to learn a new clause */\\
					$level \leftarrow$ the second highest level in $newLearntClause$;\\
					backtrackTo$(level)$; /* cancel all variable assignments higher than level $level$ */\\
					$currentLevel \leftarrow level$;
				}
			}
			\Else{
				\If {all variables are assigned}{
					return SATISFIABLE;\\
				}
				\ElseIf{restart condition is satisfied} {
					backtrackTo(0); /* cancel all assignments depending on a decision */\\
					break; /* restart */
				}
				\ElseIf {learnt clause database reduction condition is satisfied} {
					remove a subset of learnt clauses selected using a heuristic;
				} 
				\Else{
				        $currentLevel++$;\\
					$x \leftarrow$ a non-assigned variable selected using some heuristic;\\
					add the unit clause $x$ or $\neg x$ into $\phi$ according to a polarity heuristic such as phase saving;
				}
			}
		}
	}
}
\caption{CDCL$(\phi)$, a generic CDCL SAT algorithm \label{CDCL}}
\end{algorithm}

 \begin{algorithm}[!tb]
\KwIn{$\phi$: A CNF formula with original and learnt clauses;  $nbNewLearnts$: the number of clauses learnt since the last execution of function vivify$(\phi)$; $\sigma$: the number of times function vivify$(\phi)$ was executed so far.}
\KwOut{$\phi$ with some simplified clauses}
\Begin{
	\If {liveRestart$(nbNewLearnts, \sigma)$} {
	   $\phi \leftarrow$ vivify$(\phi)$; /* call Algorithm~\ref{vivifying} */\\
	   $nbNewLearnts \leftarrow 0$; $\sigma++$;
	}
	return $\phi$;
}
\caption{vivifyIfPromising$(\phi, nbNewLearnts, \sigma)$, activate clause vivification if it is judged to be promising \label{minimization}}
\end{algorithm}

Given a clause $C= l_1 \vee l_2 \vee \cdots \vee l_k$ such that liveClause$(C)$ is true, Algorithm \ref{vivifying} (i.e., function vivify$(\phi)$) applies the following simplification rules, using the same data structure and unit propagation algorithm as in the CDCL SAT solver:

\begin{enumerate}
\item  {\bf Rule\_1:} If UP$(\phi \cup \{\neg l_1, \ldots, \neg l_{i-1}\})$ deduces $\neg l_i$ for some $i\leq k$, then clause $l_1\vee \cdots \vee l_{i-1} \vee \neg l_i$ is a logical consequence of $\phi$. The algorithm removes $l_i$ from $C$  and tries to further vivify $l_1 \vee \cdots \vee  l_{i-1} \vee  l_{i+1} \vee \cdots \vee l_k$.

\item {\bf Rule\_2:} If UP$(\phi \cup \{\neg l_1, \ldots, \neg l_{i-1}\})$ deduces $l_i$ for some $i\leq k$, then clause $l_1\vee \cdots \vee l_{i-1} \vee l_i$ is a logical consequence of $\phi$. 
Hence, $C$ could be replaced with clause $l_1 \vee \cdots \vee  l_{i-1} \vee l_{i}$ because it subsumes $C$. However, the algorithm replaces $C$ with a clause obtained after analyzing the implication graph that allows to derive $l_i$. Let $R$ be the set of literals of the reason clause of $l_i$. Note that all literals in $R$ but $l_i$ are false. The algorithm replaces $l_i$ with $\neg l_i$ in $R$ to obtain a set $R$ in which  all literals are false. Then, the algorithm executes function conflAnalysis$(\phi,  \neg C' \cup \{\neg l_i\}, R)$, where $\neg C' = \{\neg l'_1,  \neg l'_2, \ldots, \neg l'_{i-1}\} \subseteq \{\neg l_1,  \neg l_2, \ldots, \neg l_k\}$  is a subset of negated literals of $C$ already propagated,  that retraces the implication graph from the literals in $R$ until the literals of $\neg C' \cup \{\neg l_i\}$, in order to collect the literals of $\neg C' \cup \{\neg l_i\}$ from which there is a path to a literal of $R$ in the implication graph. The function returns a disjunction of the negation of the collected literals.

Algorithm~\ref{conflAnalysis} implements function conflAnalysis$(\phi,  \neg C' \cup \{\neg l_i\}, R)$. Note that literals in $\neg C' \cup \{\neg l_i\}$ do not have any reason clause, but other asserted literals do. A literal $l$ of  $\neg C' \cup \{\neg l_i\}$ such that $seen[l]==0$ in line \ref{lineSeen} is a literal from which no path exists to the conflict represented by $R$ and will not be collected in $D'$. For example, in Figure~\ref{implicationFig}, $\neg C' \cup \{\neg l_i\}=\{l_1, l_{8}, l_{20}, l_{30}\}$, there is no path from $l_{20}$ to $\Box$, and function conflAnalysis$(\phi,  \neg C' \cup \{\neg l_i\}, R)$ returns $\neg l_{1} \vee \neg l_{8} \vee \neg l_{30}$.


\item If UP$(\phi  \cup \{\neg l_1, \ldots, \neg l_{i-1}\})$ neither deduces $l_i$ nor  $\neg l_i$. We distinguish two cases:
\begin{enumerate}
\item {\bf Rule\_3:} If UP$(\phi \cup \{\neg l_1, \ldots, \neg l_i\})=\Box$, then $\phi \cup \{\neg l_1,  \ldots, \neg l_i\}$ is unsatisfiable and clause $l_1\vee\cdots\vee l_i$ is a logical consequence of $\phi$ and could replace $C$. 
However, as before, let $R$ be the set of literals of the falsified clause, the algorithm replaces $C$ with a disjunction of a subset of negated literals of $\neg C' \cup \{\neg l_i\}$ returned by function $conflAnalysis(\phi,  \neg C' \cup \{\neg l_i\}, R)$, which is a sub-clause of $l_1\vee\cdots\vee l_i$.

\item {\bf Rule\_4:} If UP$(\phi \cup \{\neg l_1, \ldots, \neg l_i\}) \neq \Box$, then the literal $l_i$ is added to the working clause $C'$, and the algorithm continues processing the literal $l_{i+1}$.
\end{enumerate}
\end{enumerate}

Observe that Rule\_2 and Rule\_3 cannot be both applied to a clause $C$ during the execution of Algorithm~~\ref{vivifying}, whereas Rule\_1 and Rule\_4 can be combined with Rule\_2 and Rule\_3.

\begin{algorithm}[!tb]
\KwIn{$\phi$: A CNF formula}
\KwOut{$\phi$ with some simplified clauses}
\Begin{
	\ForEach {$C=l_1 \vee \cdots \vee l_k \in \phi$}  {
		{\bf if} { $!liveClause(C)$} {\bf then} {continue;}\\
		$C \leftarrow sort(C)$; 
		$C' \leftarrow \emptyset$; /* $C'$ will be the vivified clause */\\
		\For {$i:=1\ to\ k$} {
		        {\bf if} { $l_i$ is false} {\bf then} {continue;} /* Rule 1 */\\
		        \If{$l_i$ is true}{ 
		        	     Copy the literals of the reason clause of $l_i$ into $R$;\\
			     $R \leftarrow (R\setminus \{l_i\}) \cup \{\neg l_i\}$;\\
		              $C'  \leftarrow conflAnalysis(\phi, \neg C'\cup \{\neg l_i\}, R)$; /* Rule 2: call Algorithm~\ref{conflAnalysis} */\\
			     break;
			   }
		        \ElseIf {$(R\leftarrow$ UP$(\phi \cup \neg C' \cup \{\neg l_i\}))$ is a falsified clause \label{up1}} {
			    $C'  \leftarrow conflAnalysis(\phi, \neg C'\cup \{\neg l_i\}, R)$; 
			    /* Rule 3: call Algorithm~\ref{conflAnalysis} */\\
			     break;
			}
			\Else{$C'  \leftarrow C' \vee l_i$; /* Rule 4: add $l_i$ to clause $C'$ */
			}
		}
	}
	$\phi  \leftarrow  (\phi \setminus \{C\}) \cup \{C'\}$;\\
	return $\phi$;\\
}
\caption{vivify$(\phi)$: vivifying clauses in $\phi$ \label{vivifying}}
\end{algorithm}

\begin{algorithm}[!tb]
\KwIn{$\phi$: A CNF formula with original and learnt clauses;   $D$: a subset of negated literals of a clause 
that are propagated by vivification; $R$: a set of literals falsified by vivification, representing a conflict.}
\KwOut{a disjunction $D'$ of a subset of negated literals of $D$}
\Begin{
	$D' \leftarrow \emptyset$;\\
	{\bf foreach} {\em literal $l$ in $R$} {\bf do} $seen[l] \leftarrow 1$;\\
	\ForEach {asserted literal $l$ in the decreasing order of their asserted time}  {
		\If{$seen[l] == 1$}{ \label{lineSeen}
			\If{$l \in D$}{
				$D'  \leftarrow D' \vee \neg l$; /* a path exists from $l$ to a literal in $R$: $l$ contributes to the conflict $R$ */
			}
			\Else {
				$Reason \leftarrow$ the reason clause of $l$;\\
				{\bf foreach} {\em literal $l'$ in $Reason$} {\bf do} $seen[l'] \leftarrow 1$;\\
			}
		}
	}
   {\bf foreach} {\em literal $l$ such that {$seen[l] == 1$} } {\bf do}
   		$seen[l] \leftarrow 0$;\\
	return $D'$;\\
}
\caption{conflAnalysis$(\phi,  D, R)$: inspecting the complete implication graph to collect a disjunction of a subset of negated literals of $D$ \label{conflAnalysis}}
\end{algorithm}

The vivification function vivify$(\phi)$ proposed in this paper is a variant of the vivification function of Piette et al.\  implemented in the ReVivAl preprocessor~\cite{PHS08}. The differences between the two functions mainly come from the fact that the vivification function of Piette et al.\ is proposed for preprocessing while our function is designed for inprocessing (Piette et al.\ only mention inprocessing  vivification application as future work). These differences are indeed decisive for obtaining an effective and efficient inprocessing approach and could be stated as follows:

\begin{itemize}
\item The function of Piette et al.\ vivifies all original clauses and rechecks previously checked clauses when a redundant literal is removed
, while our function vivifies only a selected subset of original and learnt clauses and does not re-check previously vivified clauses. In fact, re-checking previously vivified clauses can be done in preprocessing but would be too costly during inprocessing.

\item The approach of Piette et al.\ orders the literals of clauses using a MOMS-style heuristic, while our approach will use the existing order of literals in the clauses. In fact, as we will explain in the next subsection, the two-literal watching technique in modern CDCL SAT solvers continuously changes the order of the literals in a clause during the search, possibly favouring the success of the vivification function, so that inprocessing vivification does not need a MOMS-style heuristic to re-order the literals of a clause. 

\item The third rule of Piette et al.\ uses the first UIP scheme to extract a nogood, while our Rule 2 and Rule 3 derive a sub-clause by inspecting the complete implication graph. In  preprocessing, one can choose between using the first UIP scheme or inspect the complete implication graph. Piette et al. mention the idea of inspecting the complete implication graph but did not apply it for efficiency reason. However, during inprocessing, one must inspect the complete implication graph to generate a sub-clause of a clause $C$ to substitute $C$.
In fact, the purpose of inprocessing vivification is to simplify the formula, but the first UIP scheme gives a nogood which often is not a sub-clause of $C$ and has to be added as a redundant clause, increasing the size of the formula.

\item The first rule of Piette et al.\ does not use any conflict analysis to extract an even smaller sub-clause, which may not be a problem in preprocessing because $C$ can be re-vivified. In the same situation, our Rule 2 inspects the complete implication graph to generate a sub-clause as small as possible. This avoids to re-vivify the same clause during inprocessing, where the cost  is significant.
\end{itemize}

The implementation of Algorithm~\ref{vivifying} in a particular solver needs the definition of the functions $liveClause(C)$ and $sort(C)$ for that solver. We will define functions $liveRestart(nbNewLearnts, \sigma)$ and $liveClause(C)$ for five of the best performing state-of-the-art solvers. A constraint we impose for now is that $liveClause(C)$ can be true only if $C$ is a learnt clause that was never  vivified before. This constraint will be later removed in Section~\ref{approachImprovement}, where it will be allowed to vivify both original and learnt clauses, as well as re-vivify clauses under some conditions. As for function sort$(C)$,  it simply returns $C$ in its current literal order. The next subsection analyzes the reason of this choice.

\subsection{The Current Literal Order in a Clause \label{SectOrder}}

When a SAT solver derives a conflict in the current level, it derives a learnt clause by retracing the implication graph from the conflict until a literal of a lower level or the first UIP in each path.  In state-of-the-art solvers such as MiniSat and its descendants, the implication graph is retraced from the conflict in a breadth-first manner, so that the literals are put in the learnt clause in increasing order of their distance to the conflict, except the two first literals that will be watched after backtracking: the first literal should be the negation of the first UIP and the second literal should be a literal of the second highest level in the learnt clause.
For example, the learnt clause derived from the implication graph in Figure \ref{implicationFig} is $\neg l_{45} \vee \neg l_8 \vee \neg l_5 \vee \neg l_{16} \vee \neg l_{11}$ (in this order).


Detecting a redundant literal in a clause $C$ using Algorithm \ref{vivifying} means that Algorithm \ref{vivifying} can derive a conflict without needing to propagate the negation of all literals of $C$. This happens when propagating the negation of a subset of literals of $C$ already imply the literals in the paths from the negation of the other literals of $C$ to the conflict. For example, in the learnt clause $\neg l_{45} \vee \neg l_8 \vee \neg l_5 \vee \neg l_{16} \vee \neg l_{11}$ derived from Figure \ref{implicationFig}, if propagating $l_{45}$ and $l_8$ already asserts $l_{140}$ or $l_{100}$, which are in the path from $l_{16}$ to the conflict, then clause vivification detects that $\neg l_{16}$ is redundant in the learnt clause. 

Consequently, a reasonable hypothesis is that propagating in priority literals closer to the conflict allows to detect more easily redundant literals. In fact, other things being equal,  the longer is a path, the higher is the probability that the path contains a literal that can be asserted by propagating other literals.
 
Therefore, the original literal order of a learnt clause $C$, which roughly is in the increasing order of their distance to the conflict, is suitable for Algorithm \ref{vivifying} to detect redundant literals.

Nevertheless, the literal order of a (learnt or original) clause is continuously changed during the search, if the solver uses the two-literal watching technique and always watches the first two literals of the clause. We argue that these changes increase the probability that Algorithm \ref{vivifying} detects a redundant literal. To see this, we present below how the changes are made in a clause $C$ during the search when one of the first two literals in $C$ is falsified (no change is made in $C$ in any other case). 

\begin{enumerate}
\item If the other watched literal is satisfied, no change is made in $C$;

\item Otherwise, let $C=\l_1 \vee \l_2 \vee \l_3 \vee \cdots \vee l_k$. 
The falsified literal is placed to be $l_2$ (by exchanging with $l_1$ if necessary). 
If $C$ becomes unit or empty, no further change is made on $C$. Otherwise,
assume that $l_i$ ($i>2$) is the first non-falsified literal of $C$. Literals $l_2$ and $l_i$ are exchanged so that $l_i$ is watched in the place of $l_2$.  After the exchange, $C$ becomes $\l_1 \vee \l_i \vee \l_3 \vee \cdots \vee l_{i-1} \vee l_2 \vee l_{i+1} \vee \cdots \vee l_k$. We can see that literals $l_3, l_4, \ldots, l_{i-1}, l_2$ are falsified.  If the falsification of $l_3, l_4, \ldots, l_{i-1}, l_2$ does not allow to derive a conflict,  the search continues and the next non-falsified literals of $C$ can be pushed ahead. The solver stops changing the literal order of $C$ when a conflict is derived or one of the two first literals is satisfied.
\end{enumerate}

In summary, the two-literal watching technique, applied during the search and clause vivification, changes the literal order of a clause $C$ by pushing the falsified literals not allowing to derive a conflict to the end of $C$ and by placing more promising literals at the beginning of $C$. Recall that clause vivification is based on successively falsifying the literals of $C$ to derive a conflict. The current literal order obtained after the search presumably increases the success probability of vivifying $C$. That is why function sort$(C)$ just returns $C$ in its current literal order in our approach. In Section \ref{experiments}, we will present experimental results to show that the current literal order is indeed better than several other orders.

\subsection{Vivifying Learnt Clauses in Glucose and TC\underline{\hspace{2mm}}Glucose}

Glucose is a very efficient CDCL SAT solver developed from MiniSat \cite{een2003extensible}. It was habitually awarded in the SAT Competition between 2009 and 2014, and is the base solver of many other awarded solvers.

Glucose was the first solver which incorporated the LBD measure in the clause learning mechanism and adopted an aggressive  strategy for clause database reduction. We used Glucose 3.0, in which the reduction process is fired once the number of clauses learnt since the last reduction reaches $first + 2\times inc \times \sigma$, where $first=2000$ and $inc=300$ are parameters (note that $2\times inc$ can be considered as a single parameter in Glucose 3.0), and $\sigma$ is the number of database reductions performed so far. The learnt clauses are first sorted in decreasing order of their LBD values, and then the first half of learnt clauses are removed except for the binary clauses, the clauses whose LBD value is 2 and the clauses that are reasons of the current partial assignment. Note that the reduction process is not necessarily fired at level 0.

Glucose 3.0 also features a fast restart mechanism which is independent of the clause database reduction. Roughly speaking, Glucose restarts the search from level 0 when the average LBD value in recent learnt clauses is high compared with the average LBD value of all the learnt clauses.

Solver TC\underline{\hspace{2mm}}Glucose is like Glucose 3.0 but it uses a tie-breaking technique for VSIDS,  and the CHB branching heuristic~\cite{LGPC16aaai} instead of the VSIDS branching heuristic for small instances. It was the best solver of the hard combinatorial category in SAT Competition 2016.

Learnt clause vivification in Glucose and TC\underline{\hspace{2mm}}Glucose is implemented by defining the three functions in Algorithm~\ref{minimization} and~\ref{vivifying} as follows:

Function $liveRestart(nbNewLearnts, \sigma)$ returns true iff the learnt clause reduction process was fired in the preceding restart. In other words, clause vivification in Glucose and  TC\underline{\hspace{2mm}}Glucose follows their learnt clause database reduction. 

Function $liveClause(C)$ returns true iff $C$ is a learnt clause that has not yet been vivified and belongs to the second half of learnt clauses after sorting the clauses in decreasing order of their LBD values. 

Function $sort(C)$ returns $C$ without changing the order of its literals.

The intuition behind the definition of $liveRestart(nbNewLearnts, \sigma)$ and $liveClause(C)$ for Glucose can be stated as follows. Just after the learnt clause database reduction, about a half of learnt clauses remain there. Among these remaining learnt clauses, the half with smaller LBD will probably survive the next learnt clause database reduction. The vivification of this half of remaining learnt clauses (i.e., a quarter of all learnt clauses) is useful and has a moderate cost. 


\subsection{Vivifying Learnt Clauses in COMiniSatPS, MapleCOMSPS and MapleCOMSPS\underline{\hspace{2mm}}LRB}\label{maple-min}

COMiniSatPS is a SAT solver created by applying a series of small diff patches to MiniSat 2.2.0. Its  initial prototypes (SWDiA5BY and MiniSat\underline{\hspace{2mm}}HACK\underline{\hspace{2mm}}xxxED) won six medals in SAT Competition 2014 and Configurable SAT Solver Challenge 2014.
MapleCOMPS and MapleCOMSPS\underline{\hspace{2mm}}LRB are based on COMiniSatPS, and were the winners of the main track and the application category in SAT Competition 2016, respectively.

The clause database reduction policy of COMiniSatPS, MapleCOMSPS and MapleCOMSPS\underline{\hspace{2mm}}LRB is quite different from that of Glucose. In these solvers, the learnt clauses are divided into three subsets: (1) clauses whose LBD value is smaller than or equal to a threshold $t_1$ are stored in a subset called $CORE$; (2) clauses whose LBD value is greater than $t_1$ and smaller than or equal to another threshold $t_2$ are stored in a subset called $TIER2$; and (3) the remaining clauses are stored in a subset called $LOCAL$. If a clause in $TIER2$ is not involved in any conflict for a long time, it is moved to $LOCAL$.

Periodically, the clauses of $LOCAL$ are sorted in increasing order of their activity in recent conflicts, and the learnt clauses in the first half are removed (except for the clauses that are reasons of the current partial assignment).

The three solvers interleave Glucose-style restart phases with phases without restarts and Luby restart phases.
In a Glucose-style restart phase, search is restarted from level 0 if the average LBD
value of recent learnt clauses is high. In a Luby restart phase, search is restarted
after reaching a number of conflicts. This number can be small (in this case, the restart is fast), and can also be high (in this case, the restart is long).

Clause vivification in COMiniSatPS, MapleCOMSPS and MapleCOMSPS\underline{\hspace{2mm}}LRB is implemented by defining the three functions in Algorithm \ref{minimization} and Algorithm \ref{vivifying} as follows (recall that Algorithm \ref{minimization} is executed before each restart):

Function $liveRestart(nbNewLearnts, \sigma)$ returns true iff $nbNewLearnts$, the number of clauses learnt since the last learnt clause vivification, is greater than or equal to $\alpha + \beta\times \sigma$. We empirically fixed $\alpha$=1000 and $\beta$=2000 for the three solvers. Note that function $liveRestart(nbNewLearnts, \sigma)$ does not follow the clause database reduction in any of the three solvers.

Function $liveClause(C)$ returns true iff $C$ is a learnt clause that has not yet been vivified and belongs to $CORE$ or $TIER2$.

Function $sort(C)$ returns $C$ without changing the order of its literals.

The definition of $liveRestart(nbNewLearnts, \sigma)$ in COMiniSatPS, MapleCOMSPS and MapleCOMSPS\underline{\hspace{2mm}}LRB is inspired by the learnt clause database reduction of Glucose, because Glucose is the first solver in which we have made effective our inprocessing clause vivification approach by following its learnt clause database reduction strategy. 
In the definition of  $liveRestart(nbNewLearnts, \sigma)$, $\alpha + \beta\times \sigma$ essentially imposes an interval between two successive clause vivifications, and the length of this interval is growing by a constant $\beta$: Let $k$ be the length of the interval between the $(i-1)^{th}$ and the $i^{th}$ clause vivifications, then the length of the interval between the $i^{th}$ and the $(i+1)^{th}$ clause vivifications is $k+\beta$. 
The original intention of this definition of $liveRestart(nbNewLearnts, \sigma)$ is to vivify learnt clauses more frequently at the beginning of the search and to reduce the vivification frequency gradually as the search proceeds, because the quality of learnt clauses is generally lower at the beginning of the search.

\section{Experimental Investigation of Learnt Clause Vivification} \label{experiments}

We implemented the learnt clause vivification approach described in Section~\ref{mainSection} in the solvers Glucose 3.0, TC\underline{\hspace{2mm}}Glucose, COMiniSatPS (COMSPS for short), MapleCOMSPS (Maple for short) and MapleCOMSPS\underline{\hspace{2mm}}LRB (MapleLRB for short). The resulting solvers\footnote{Available at http:\hspace{-0.1mm}//home.mis.u-picardie.fr/\~{}cli/} are named
Glucose+, TC\underline{\hspace{2mm}}Glucose+, COMSPS+, Maple+ and MapleLRB+, respectively. Besides, we created the solvers MapleLRB/noSp and  MapleLRB+/noSp by disabling in MapleLBR and MapleLRB+, respectively, the inprocessing technique of~\cite{heule2011efficient} that  implements function $Stamp$ and is based on binary implication graphs (noSp means that function $Stamp$ is removed in MapleLRB/noSp and  MapleLRB+/noSp). 

The test instances include the application and hard combinatorial tracks of SAT Competition 2014 and 2016 
and the instances from the main track of SAT Competition 2017 (no distinction is made between the application and hard combinatorial instances in this track in the 2017 edition).
The experiments reported in this section were performed on a computer with Intel Westmere Xeon E7-8837 processors at 2.66 GHz and 10 GB of memory under Linux unless otherwise stated. The cutoff time is 5000 seconds for each instance and solver.

\subsection{Effectiveness of the learnt clause vivification approach}

Table~\ref{comparisonApp} compares the solvers Glucose, Glucose+, COMSPS, COMSPS+, Maple, Maple+, MapleLRB, MapleLRB+, MapleLRB/noSp and MapleLRB+/noSp on instances of the application category of SAT Competition 2014 and 2016. The proposed learnt clause vivification approach consistently improves the performance of all the solvers. In particular, MapleLRB was the best solver in this category in 2016, but it solved only 4 instances more than the 4th and 5th solvers of this category. However, learnt clause vivification allows MapleLRB+ to solve 15 instances from 2016 more than MapleLRB. 

\begin{table}
\centering
\begin{tabular}{|l|c|c|c|c|c|c|}
\hline
& \multicolumn{3}{c}{SAT 2014 (300 instances)} & \multicolumn{3}{|c|}{SAT 2016 (300 instances)}\\
\hline
Solver & total & Sat\  & Unsat\  & total & Sat\   & Unsat\  \\
\hline
Glucose & 213 & 100 & 113 & 146 & 61 & 85 \\
Glucose+ & 224 & 105 & 119 & 152 & 63 & 89 \\
COMSPS & 221 & 105 & 116 & 146 & 65 & 81\\
COMSPS+ & 236 & 112 & 124 & 155 & 66 & 89\\
Maple         & 235 & 114 & 121 & 154 & 72 & 82\\
Maple+       & 250 & 117 & 133 & 161 & 70 & 91\\
MapleLRB & 234 & 111 & 123 & 152 & 68 & 84\\
MapleLRB+ & 248 & 114 & 134 & 167 & 74 & 93\\
MapleLRB/noSp & 231 & 111 & 120 & 150 & 67 & 83\\
MapleLRB+/noSp & 247 & 113 & 134 & 164 & 73 & 91\\
\hline
\end{tabular}
\caption{\small {Number of solved instances (Total, Sat,  Unsat) of Glucose, Glucose+, COMSPS, COMSPS+, Maple, Maple+, MapleLRB and MapleLRB+, MapleLRB/noSp and MapleLRB+/noSp on application instances of SAT Competition 2014 and 2016.}}
\label{comparisonApp}
\end{table}

\begin{figure}
 \centering

   \begin{minipage}{0.95\linewidth}
    \centering \small ak* family benchmarks (20 instances, of which 13 unsolved
instances are at point(5000, 5000)). MapleCOMSPS\_LRB+
solves 6 instances more than MapleCOMSPS\_LRB
(represented by the points $(x, 5000)$ where $x<5000$).
  
  \centering
    \includegraphics[width=0.4\textwidth]{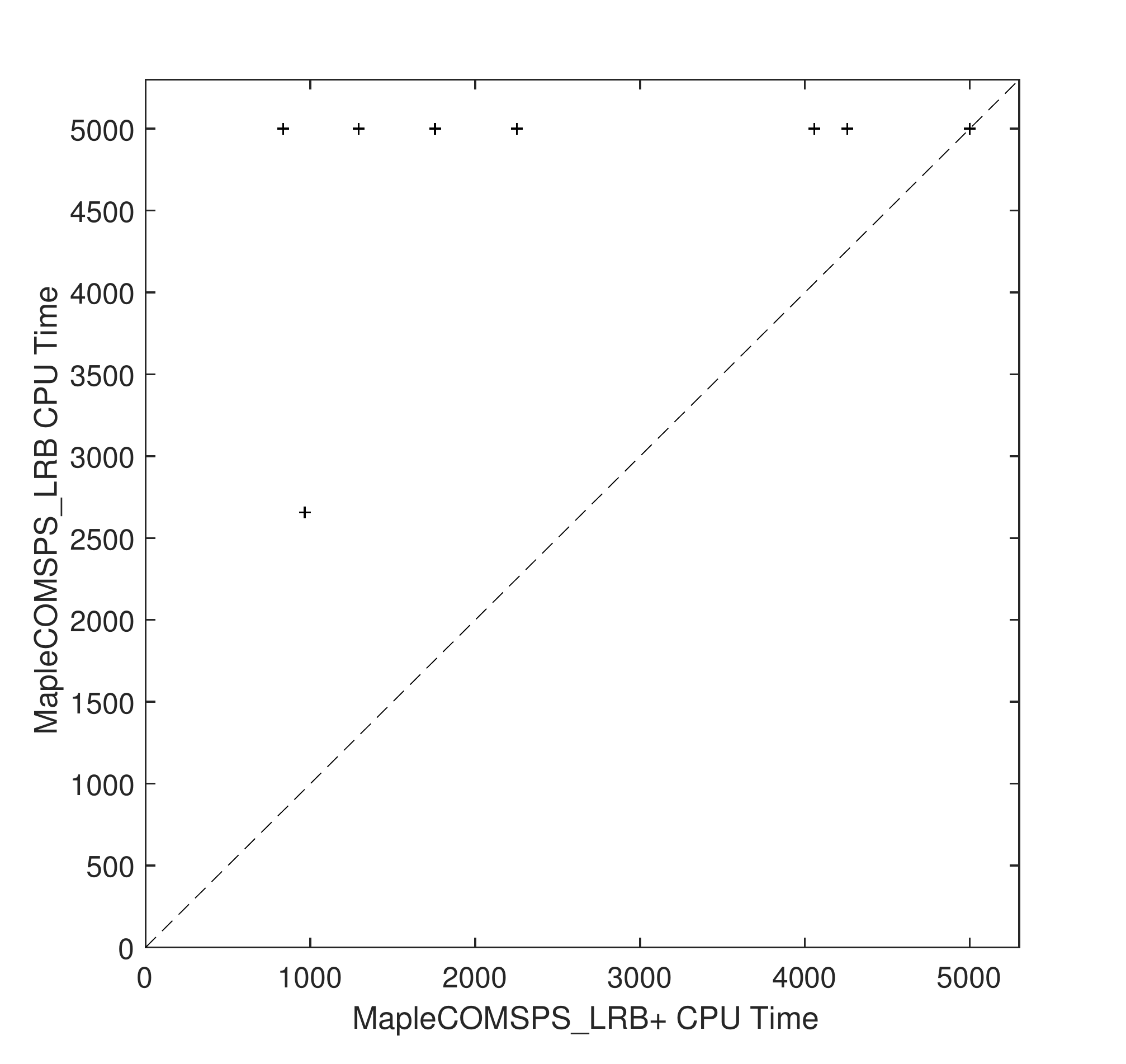}
    \label{figure_ak-MapleLRB}
  \end{minipage}

  \begin{minipage}{0.95\linewidth}
 \centering \small sokoba* family benchmarks(33 instances, of which 13 unsolved instances are at point(5000, 5000)). MapleCOMSPS\_LRB+
solves 5 instances more than MapleCOMSPS\_LRB (represented by the points $(x, 5000)$ where $x<5000$).

    \centering
    \includegraphics[width=0.4\textwidth]{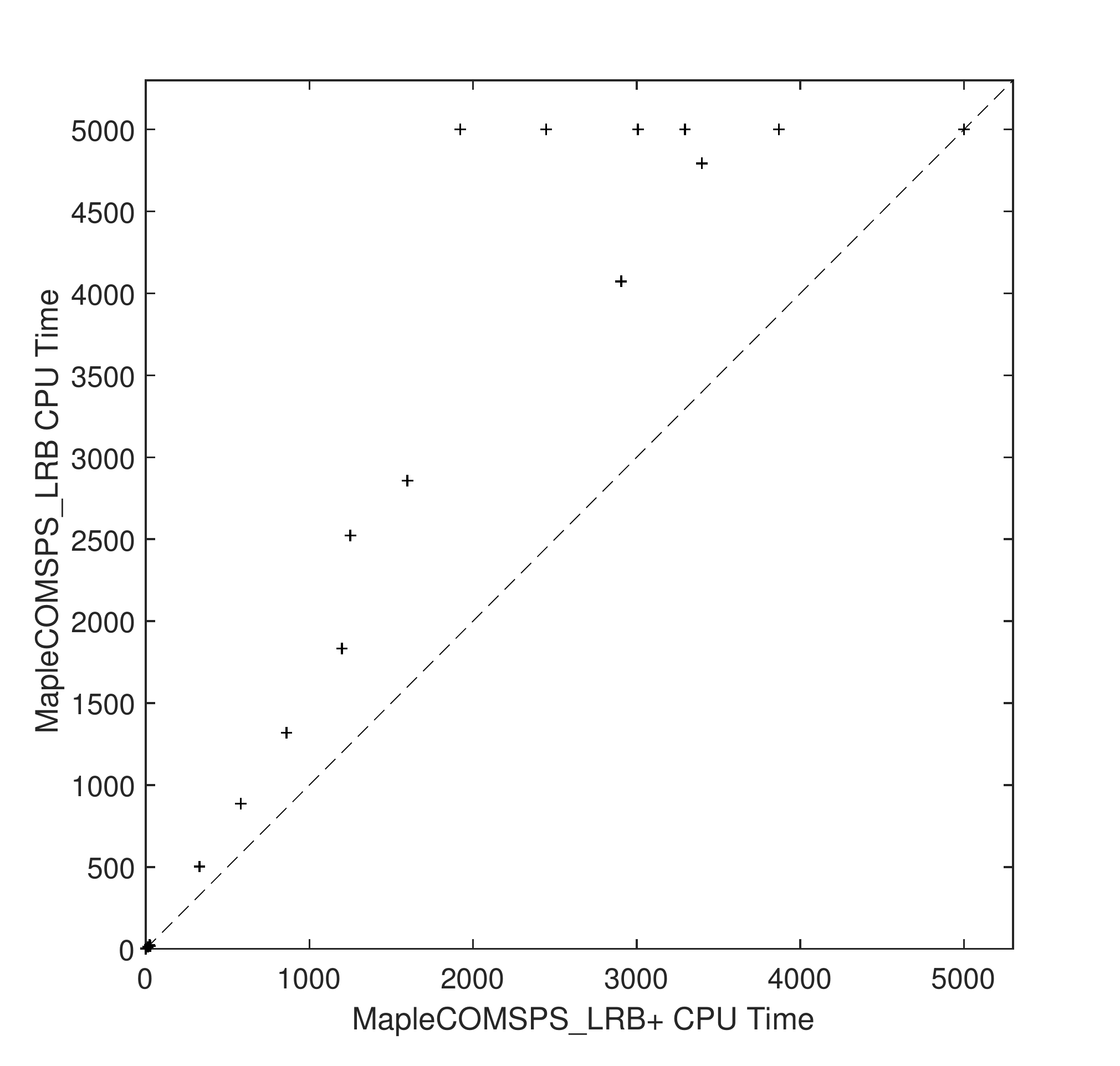}
    \label{figure_sokoban-MapleLRB}
  \end{minipage}%
  
   \begin{minipage}{0.95\linewidth}
    \centering \small Sz* family benchmarks(8 instances, of which 4 unsolved
instances are at point(5000, 5000)). MapleCOMSPS\_LRB+
solves 3 instances more than MapleCOMSPS\_LRB
(represented by the points $(x, 5000)$ where $x<5000)$.
    
     \centering
    \includegraphics[width=0.4\textwidth]{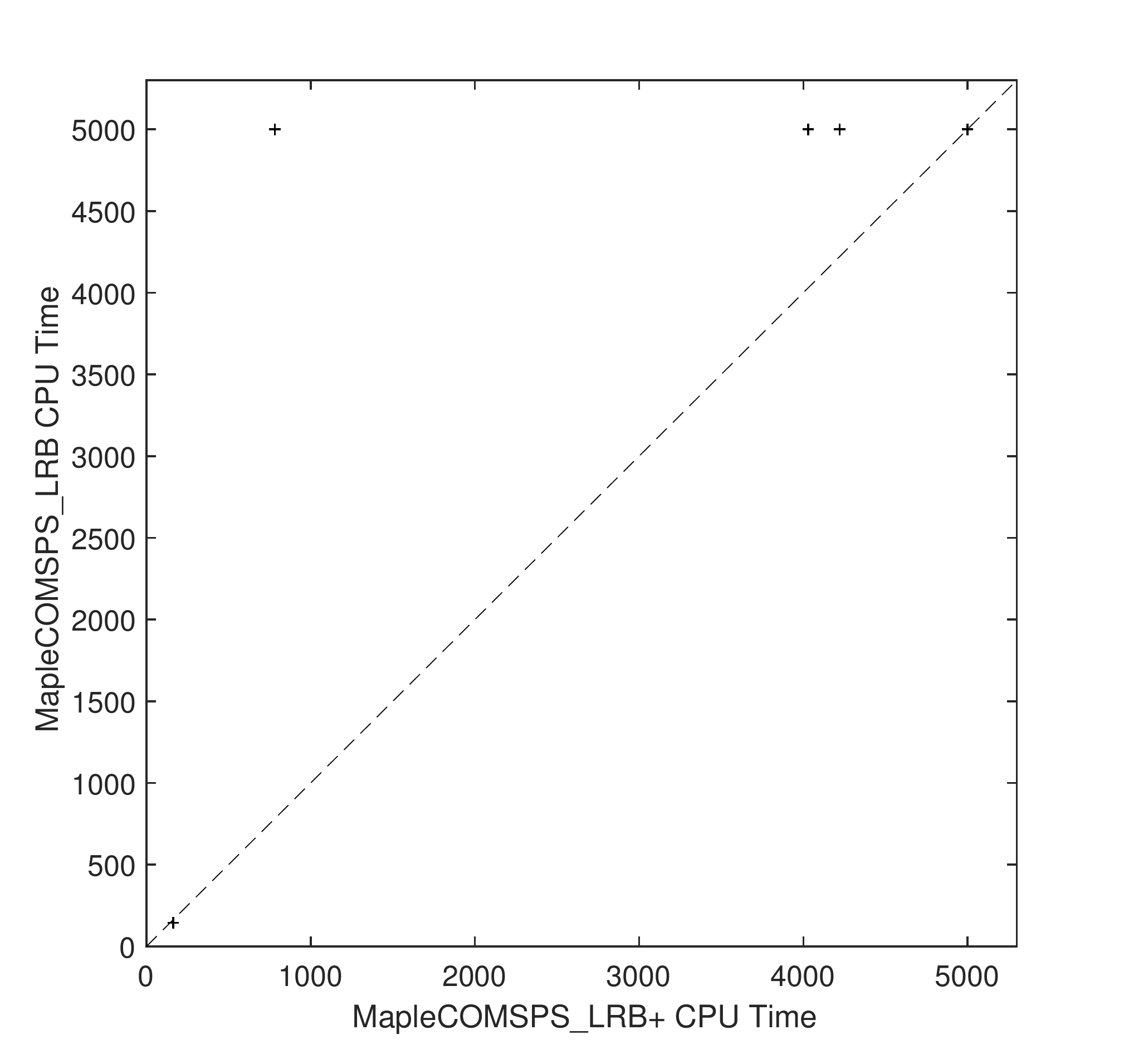}
    \label{figure_Sz-MapleLRB}
  \end{minipage}
\caption{\small Scatter plots comparing MapleLRB+ runtime ($x$-axis) and MapleLRB runtime ($y$-axis) on 3 benchmark families of the application category of SAT Competition 2016. MapleLRB+ solves 14 instances more in these three families.}
\label{figure_family}
\end{figure}

Figure~\ref{figure_family} shows a sample of scatter plots comparing MapleLRB+
and MapleLRB on 3 benchmark families of the application
category of SAT Competition 2016, where a family consists of instances with similar names.
A point $(x, y)$ in the plots corresponds to an instance, where $x$ ($y$) represents the solving time in seconds of
MapleLRB+ (MapleLRB). A point $(x, y)$ where $x=5000$ ($y=5000$) means that
the instance was not solved  by MapleLRB+ (MapleLRB) within a cutoff time of 5000s.

The performance of MapleLRB+ and MapleLRB is similar on other benchmark families of the application
category of SAT Competition 2016. The difference in the number of solved instances within the cutoff time between the two solvers is smaller than or equal to 1.

Table~\ref{comparisonApp} also compares the proposed inprocessing vivification approach with the inprocessing of~\cite{heule2011efficient} using four solvers: MapleLRB, MapleLRB+, MapleLRB/noSp and MapleLRB+/noSp. 
The differences among these four solvers are specified in their names: 

\begin{description}
\item {\bf MapleLRB+:}  with the proposed vivification and the inprocessing of~\cite{heule2011efficient} (implemented in function Stamp).
\item {\bf MapleLRB+/noSp:}  with the proposed vivification but without Stamp.
\item {\bf MapleLRB:}  without the proposed vivification  but with Stamp.
\item {\bf MapleLRB/noSp:}  without the proposed vivification and without Stamp.
\end{description}

The inprocessing of~\cite{heule2011efficient} implemented in MapleLRB allows MapleLRB to solve 3 (2) instances of 2014 (2016) more than MapleLRB/noSp, which has  disabled that inprocessing. However, our approach is more effective, allowing MapleLRB+/noSp to solve 16 (14)  instances of 2014 (2016) more than MapleLRB/noSp, and MapleLRB+ to solve 14 (15) instances of 2014 (2016) more than MapleLRB. 

\begin{figure}[!h]
\centering
\begin{minipage}{0.8\linewidth}
    \centering
    \includegraphics[width=\textwidth]{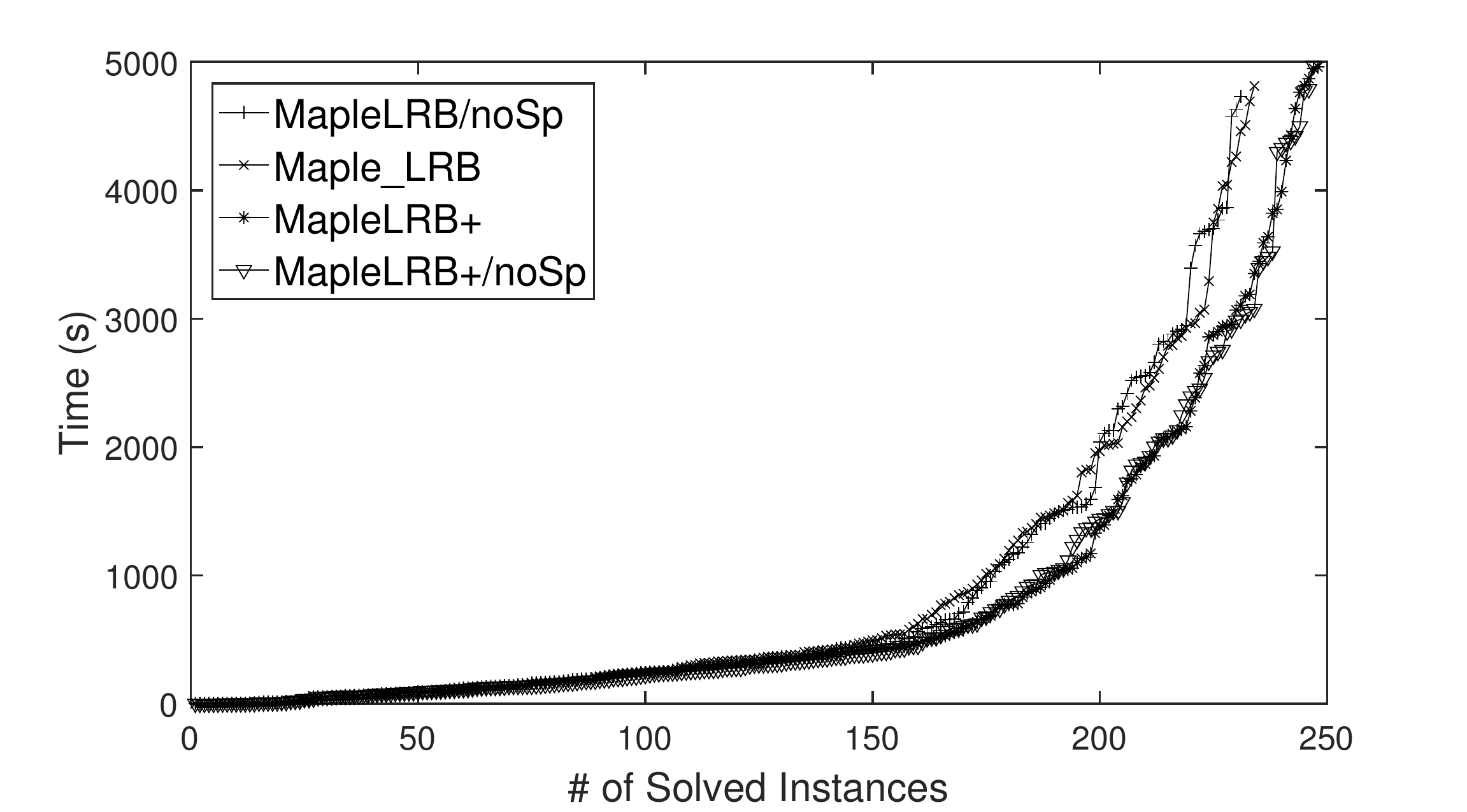}
    \label{figure_sc14}
  \end{minipage}
  \begin{minipage}{0.8\linewidth}
    \centering
    \includegraphics[width=\textwidth]{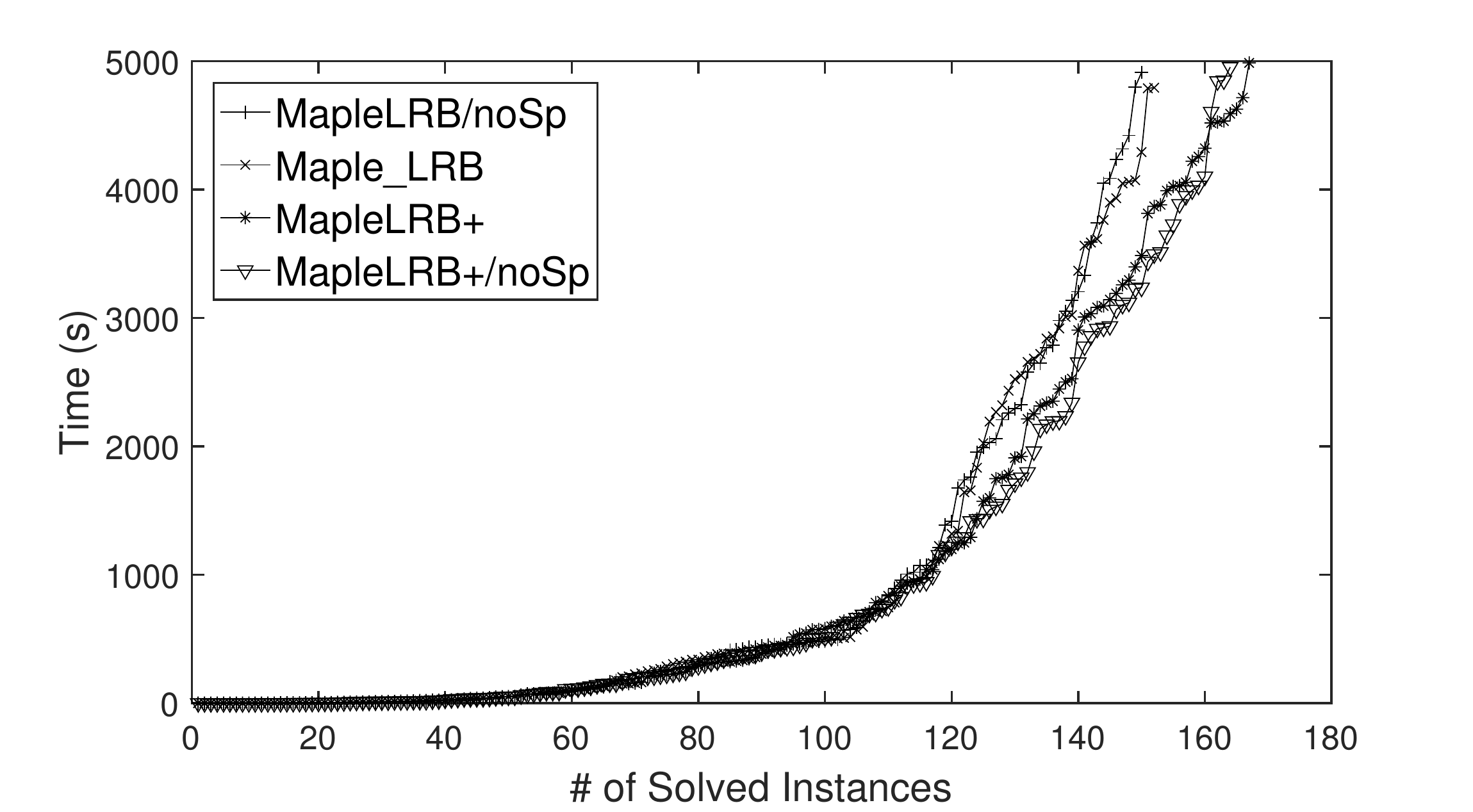}
    \label{figure_sc16}
\end{minipage}
\caption{\small Cactus plots of MapleLRB+, MapleLRB, MapleLRB+/noSP and MapleLRB/noSp on application instances of SAT Competition 2014 (top) and 2016 (bottom).}
\label{figure_cactus}
\end{figure}

Figure~\ref{figure_cactus} shows the cactus plots of the four solvers on the application instances of SAT Competition 2014 (top) and 2016 (bottom). The two solvers using the proposed vivification approach perform clearly better than the two solvers using the inprocessing of \cite{heule2011efficient}. Note that the inprocessing of~\cite{heule2011efficient} is also used in Lingeling and subsumes (at least partly) several inprocessing techniques.

Table~\ref{comparisonCrafted} compares Maple, Maple+, TC\underline{\hspace{2mm}}Glucose, TC\underline{\hspace{2mm}}Glucose+, MapleLRB, MapleLRB+, MapleLRB/noSp and MapleLRB+/noSp on the hard combinatorial instances of SAT Competition 2014 and 2016. These instances are also called {\em crafted} instances. Note that the instances of SAT Competition 2016 are very special, because the best solver  for these instances (TC\underline{\hspace{2mm}}Glucose) performs significantly less well than the other solvers on the instances of 2014. The performance of Maple+, TC\underline{\hspace{2mm}}Glucose+ and MapleLRB+ is similar to that of Maple, TC\underline{\hspace{2mm}}Glucose and MapleLRB, respectively, on the instances of 2016. However, their performance is significantly better than that of Maple, TC\underline{\hspace{2mm}}Glucose and MapleLRB on the instances of 2014. 

\begin{table}
\centering
\begin{tabular}{|l|c|c|c|c|c|c|}
\hline
& \multicolumn{3}{c}{SAT 2014 (300 instances)} & \multicolumn{3}{|c|}{SAT 2016 (200 instances)}\\
\hline
Solver & total & Sat\  & Unsat\  & total & Sat\   & Unsat\  \\
\hline
Maple & 217 & 97 & 120 & 47 & 8 & 39 \\
Maple+ & 221 & 96 & 125 & 46 & 6 & 40 \\
TC\underline{\hspace{2mm}}Glucose & 172 & 80 & 92 & 58 & 3 & 55\\
TC\underline{\hspace{2mm}}Glucose+ & 184 & 79 & 105 & 57 & 3 & 54\\
MapleLRB & 206 & 93 & 113 & 40 & 8 & 32\\
MapleLRB+ & 212 & 95 & 117 & 44 & 9 & 35\\
MapleLRB/noSp & 206 &  92  & 114  & 40 & 8 & 32 \\
MapleLRB+/noSp & 214 &  97  & 117 & 44 &  9 & 35\\
\hline
\end{tabular}
\caption{\small Number of solved instances (Total, Sat,  Unsat) of Maple, Maple+, TC\underline{\hspace{2mm}}Glucose, TC\underline{\hspace{2mm}}Glucose+, MapleLRB, MapleLRB+, MapleLRB/noSp and MapleLRB+/noSp on hard combinatorial instances of SAT Competition 2014 and 2016.}
\label{comparisonCrafted}
\end{table}

Note that all the solvers in Table~\ref{comparisonApp} and~\ref{comparisonCrafted} implement the following inprocessing techniques: the clause minimization based on binary clause resolution of Glucose and the recursive learnt clause minimization of MiniSat. In addition, MapleLRB and MapleLRB+ also include the inprocessing of~\cite{heule2011efficient}. The previous results clearly indicate that the proposed vivification approach is compatible with all these inprocessing techniques.

\subsection{Analysis of the rules of vivification}

Table~\ref{cases} compares the percentage of clauses simplified by the simplification rules of  Section~\ref{original} implemented in Algorithm \ref{vivifying}. 
 In the table, {\em noSimp} refers to the percentage of clauses $C$ such that $liveClause(C)$ is true but no redundant literal is removed from $C$ (i.e., Algorithm \ref{vivifying} propagated the negation of the literals of $C$ but did not find redundant literals in $C$), {\em rule\_$n$}, where $n=1,2,3$, refers to the percentage of clauses in which {\em only} the simplification Rule\_$n$ was applied, and {\em rule\_$1n$} where $n=2, 3$, refers to the percentage of clauses in which {\em both} the simplification Rule\_$1$ and the simplification Rule\_$n$ were applied. Observe that {\em rule\_1}, {\em rule\_2}, {\em rule\_3}, {\em rule\_12} and {\em rule\_13} cover all the simplifications of Algorithm \ref{vivifying}. Rule\_2 and Rule\_3 cannot be both applied to a clause $C$ during the execution of Algorithm \ref{vivifying}, whereas Rule\_$1$ can be combined with Rule\_2 and Rule\_3. Table~\ref{cases} also gives results for {\em rule\_$n$+rule\_$1n$}, which refers to the sum of {\em rule\_$n$} and {\em rule\_$1n$} and shows the percentage of clauses in which Rule\_$n$  is applied with or without Rule\_$1$, and gives results for {\em rule\_$1$+rule\_$12$+rule\_$13$}, which shows the percentage of clauses in which Rule\_$1$  is applied with or without Rule\_$2$ and Rule\_$3$. All results are averaged among the solved instances in each group.
We do not give results for Rule\_4 because it does not remove any literal from $C$. The results indicate that Rule\_$1$, Rule\_$2$ and Rule\_$3$ all contribute to clause vivification. Rule\_2  and Rule\_3 contribute to clause vivification with similar percentages in most cases, and their combination with Rule\_1 improves significantly the percentage of simplified clauses. Hence, any implementation of the proposed clause vivification in a SAT solver should consider all the rules, because each rule makes substantial contributions.

\begin{table}[htp]
\centering
\begin{tabular}{|l|r|r|r|r|r|}\hline
&   \multicolumn{2}{c}{SAT 2014} & \multicolumn{2}{|c|}{SAT 2016} & SAT 2017 \\
\cline{2-5}
Solver   & application  & crafted & application & crafted  & main \\ \hline
{\bf noSimp} & 33.7\% & 41.7\% & 33.2\% & 51.3\% & 39.1\% \\ \hline
{\bf rule\_1} & 10.5\% & 6.9\% & 9.9\% & 4.8\% & 10.0\% \\ \hline
{\bf rule\_2} & 14.9\% & 15.7\% & 15.4\% &28.4\%  & 16.8\% \\ \hline
{\bf rule\_3} & 16.5\% & 14.4\% & 17.3\% &12.6\%  & 14.6\%\\ \hline
{\bf rule\_12} & 9.6\% & 10.6\% & 9.6\% & 1.2\% & 8.1\% \\ \hline
{\bf rule\_13} & 14.8\% & 10.7\% & 14.6\% & 1.7\% & 11.4\% \\ \hline
{\bf rule\_3+rule\_13} & 31.3\% & 25.1\% & 31.9\% & 14.3\% & 26.0\% \\ \hline
{\bf rule\_2+rule\_12} & 24.5\% & 26.3\% & 25.0\% & 29.6\% & 24.9\% \\ \hline
{\bf rule\_1+rule\_12+rule\_13} &34.9\% & 28.2\% & 34.1\% & 7.7\% & 29.5\% \\ \hline
\end{tabular}
\caption{\small Comparison of different simplification rules in Maple+}
\label{cases}
\end{table}


\subsection{Robustess of the learnt clause vivification approach}

Table \ref{simplifyFrequentApp} compares different implementations of learnt clause vivification in Glucose and MapleLRB. These implementations are obtained by varying the  functions $liveRestart(nbNewLearnts, \sigma)$, $liveClause(C)$ and $sort(C)$.  The tested instances are the 300 instances of the application track of SAT Competition 2014. The solvers used in the experiments are the following ones:

\begin{description}

\item[{\bf Glucose+/$\alpha$-$\beta$:}] it is like Glucose+ but does not follow the learnt clause database reduction of Glucose any more. Instead, $liveRestart(nbNewLearnts, \sigma)$ returns true iff $nbNewLearnts \ge \alpha + 2\times\beta\times \sigma$.

\item[{\bf Glucose+H\_lbd:}] it is like Glucose+ but $liveClause(C)$ is true iff $C$ is a learnt clause that was never vivified before and is in the {\em first} half  of learnt clauses when these clauses are sorted in decreasing order of their LBD value. In other words, the learnt clauses with higher LBD are vivified in Glucose+H\_lbd.

\item[{\bf Glucose+$\Delta$:}] it is like Glucose+ but $liveClause(C)$ is true iff $C$ is a learnt clause that was never vivified before and is in the last $\Delta$ fraction of learnt clauses when these clauses are sorted in decreasing order of their LBD value. Glucose+ is in fact Glucose+1/2.


\item[{\bf MapleLRB+/Core:}] it is like MapleLRB+ but it vivifies every not-yet-vivified clause in $CORE$ upon every restart. It vivifies the clauses in $TIER2$ as in MapleLRB+.

\item {\bf Low2highLevel, High2lowLevel, Low2highActivity, High2lowActivity, Random and Reverse:} all these solvers are like MapleLRB+, except that function $sort(C)$  is different. In the solver Low2highLevel (Low2highActivity), literals in $C$ are ordered from small level (activity) to high level (activity). In the solver High2lowLevel (High2lowActivity), literals in $C$ are ordered from high level (activity) to small level (activity). The level of a literal in $C$ refers to the level of its last assertion. In solver Random, literals in $C$ are randomly ordered. In solver Reverse, literals in $C$ are reversed.

\end{description}

\begin{table}
\centering
\begin{tabular}{|l|c|c|c|c|c|c|}
\hline
Solver & Total & Sat\  & Unsat\  & Impact & Cost & LiveC \\
\hline
Glucose & 213 & 100 & 113 & / & /  & / \\
Glucose+/$10^3$-$10^3$ & 216 & 96 & 120 & 27.04\% & 53.11\% & 11.79\%\\
Glucose+/$10^3$-500 & 219 & 100 & 119 & 27.01\% & 60.73\% & 13.51\% \\
Glucose+/$10^3$-300 & 215 & 100 & 115 & 26.69\% & 64.02\% & 13.98\% \\
Glucose+H\_lbd & 206 & 98  &  108 & 33.20\% & 132\% & 24.05\% \\
Glucose+ & 224 & 105 & 119 & 23.92\% & 29.90\% & 6.84\% \\
Glucose+1/3 & 219 & 101 & 118 & 22.71\% & 19.43\% & 4.58\% \\
Glucose+2/3 & 219 & 100 & 119 & 26.11\% & 42.45\% & 9.38\% \\
MapleLRB & 234 & 111 & 123 & / &  / & / \\
MapleLRB+/Core & 238 & 108 & 130 & 26.73\% & 62.01\% & 22.61\% \\
Low2highLevel & 247 & 114 & 133 & 22.76\% & 57.11\% & 21.25\% \\
High2lowLevel & 237 & 105 & 132 & 24.41\% & 56.05\% & 22.59\% \\
Low2highActivity & 241 & 112 & 129 & 21.88\% & 54.45\% & 21.84\% \\
High2lowActivity & 237 & 107 & 130 & 24.28\% & 57.84\% & 21.47\% \\
Random & 241 & 109 & 132 & 24.07\% & 60.45\% & 21.99\% \\
Reverse & 233 & 106 & 127 & 18.37\% & 53.24\% & 21.24\% \\
MapleLRB+ & 248 & 114 & 134 & 27.63\% & 58.11\% & 21.72\% \\
\hline
\end{tabular}
\caption{\small Comparison of different implementations of learnt clause vivification
in Glucose and MapleLRB on the 300 application instances of SAT Competition 2014.}
\label{simplifyFrequentApp}
\end{table}

Table~\ref{simplifyFrequentApp} shows the number of (Total, Sat, Unsat) instances solved by each solver within 5000s, and some statistics about the runtime behavior of each solver. Column 5 (Impact) gives the clause size reduction measured as $(a-b)/a\times 100$, where $a$ ($b$) is the total number of literals in the vivified clauses before (after) vivification. Column 6 (Cost) gives the cost of learnt clause vivification measured as the ratio of the total number of unit propagations performed by Algorithm \ref{vivifying} to the total number of other propagations performed during the search. Column 7 (LiveC) gives the ratio of the number of clauses vivified to the total number of learnt clauses.  All data are averaged over the solved instances among the 300 application instances of SAT Competition 2014.

Several observations can be made from Table~\ref{simplifyFrequentApp}:
\begin{itemize}
\item All the described implementations of  learnt clause vivification, except for the solvers Reverse and  Glucose+H\_lbd, improve their original solver. This provides evidence of the robustness of the proposed approach. It is not necessary to fine-tune different parameters to achieve significant gains.

\item Both MapleLRB+ and MapleLRB+/Core vivify all the clauses in $CORE$ but at different times. The superior performance of MapleLRB+ over MapleLRB+/Core might be explained as follows: When MapleLRB+ vivifies a clause in $CORE$, there are usually more learnt clauses in the clause database, allowing unit propagation to deduce more easily the empty clause. 

\item Propagating the literals of $C$ in their current order in $C$ is the best option, whereas propagating the literals in the reverse order is the worst option. Recall that the literals of a learnt clause are originally roughly in increasing order of their distance to the conflict and propagating literals in that order means to propagate in priority literals closer to the conflict. Moreover, that order can be changed during the search to favour the success of vivification. See Section \ref{SectOrder} for a detailed explanation.

\item Vivifying clauses with high LBD is very costly and useless, because Glucose+H\_lbd is significantly worse than Glucose.

\item The cost of the approach in Glucose+$\Delta$ is relatively small, because the solver just removed half of the learnt clauses when learnt clause vivification was activated, which is not the case for other solvers.

\item  Glucose+ and MapleLRB+ offer the best trade-off between cost and impact, explaining their superior performance.
\end{itemize}


Table~\ref{restartStrategy} shows the results of an experiment conducted to analyze how sensitive is the proposed vivification approach to the time at which vivification is activated. The table compares Maple+ with variants of Maple+ that implement different strategies for activating learnt clause vivification: Maple+eR activates vivification at every restart,  Maple+eReduceDB activates vivification if the clause database reduction was fired in the previous restart, and Maple+0.5kConflict, Maple+1kConflict and Maple+1.5kConflict activate vivification at a restart if the number of clauses learnt since the last vivification is greater than 500, 1000 and 1500, respectively. Note that the strategy in Maple+eR is different from the strategy in MapleLRB+/Core evaluated in Table~\ref{simplifyFrequentApp} in that MapleLRB+/Core activates vivification for clauses in CORE at every restart but does not do it for the clauses in TIER2, while Maple+eR does it for clauses in both CORE and TIER2. The experiment was performed on Intel Xeon E5-2680 v4 processors at 2.40GHz and 20GB of memory under Linux, which is faster than the machine used to obtain the results in Tables~\ref{comparisonApp}, \ref{comparisonCrafted}
and \ref{simplifyFrequentApp}, and in Figures~\ref{figure_family} and~\ref{figure_cactus}. 
So, the different values of number of instances solved by Maple+ are due to the use of different processors.

The first column of Table~\ref{restartStrategy} contains the name of the solver, the second and third columns contain the results for the application and crafted instances of SAT Competition 2014, the fourth and fifth columns contain the results for the application  and crafted instances of SAT Competition 2016, the sixth column contains 
the results for the instances of the main track of SAT Competition 2017, and the seventh column totalizes the results for all the instances. For each solver and group of instances, the results are displayed in terms of the total number of solved (satisfiable and unsatisfiable) instances within the cutoff time of 5000 seconds and the mean time needed to solve these instances, as well as the clause size reduction ratio measured as $(a-b)/a\times 100$ and averaged among the solved instances in each group, where $a$  is the total number of literals in all the clauses $C$ such that $liveClause(C)$ is true before applying clause vivification and $b$ is the total number of literals in all those clauses after applying clause vivification. 

The different values among the clause vivification strategies in Table~\ref{restartStrategy} are due to the fact that the number of learnt clauses between two consecutive clause vivifications is different in the tested solvers. We observe that the reduction ratio of the number of literals is about 20\% in most cases and the performance of the solver is significantly improved with this reduction. The small reduction ratio on the crafted instances of SAT competition 2016 explains why the performance of learnt clause vivification is not so good for these instances (see Table \ref{comparisonCrafted}).

The results for Maple+ and its variants, except for Maple+eR, are similar. The number of solved instances ranges from 930 to 938, and different solvers have slight different performances in different categories.  Maple+eR is not so competitive because the reduction ratio of the number of literals  is smaller. These results indicate that our approach is robust, because varying the activation strategy does not change so much the results. This means that the proposed vivification approach does not require to fine-tune the activation strategy. It is only necessary to select an strategy that leads to a reduction of the ratio of the number of literals  close to 20\%. Note, for example, that using the strategy  for reducing the clause database implemented in Maple allows Maple+eReduceDB to solve 935 instances without requiring to fine-tune any parameter. The most crucial aspect is to determine which clauses to vivify rather than  to determine when to vivify such clauses.

\begin{table}[!h]
\centering
\begin{tabular}{|ll|l|l|l|l|l|l|}\hline
&  & \multicolumn{2}{c}{SAT 2014} & \multicolumn{2}{|c|}{SAT 2016} & SAT 2017 & Total\\
\cline{3-7}
 &  & application  & crafted & application  & crafted & main &\\ 
 Solver          &   & 300 ins.      & 300 ins. & 300 ins.     & 200 ins. & 350 ins. & 1450 ins.\\ \hline
 

{\bf Maple+} & \#Solved & 265 & 218 & 175 & 56 & 219 & 933\\
& \#Sat + \#Unsat & 121+144 & 93+125 & 74+101& 8+48 &103+116 &\\
& Mean time & 1054s & 779s &1023s &1619s  & 878s &\\ 
& Reduction ratio& 24.66\% & 22.97\% &27.95\% &9.34\%  & 20.69\% & \\ \hline
{\bf Maple+0.5kConflict} & \#Solved & 263 & 217 & 174 & 50 & 226 & 930\\
& \#Sat + \#Unsat & 119+144 & 93+124 & 73+101& 6+44 & 108+118 &\\
& Mean time & 1025s & 755s & 999s & 1660s & 906s &\\ 
&Reduction ratio& 21.67\% & 19.80\% & 26.21\% & 5.83\% & 17.76\% &\\ \hline
{\bf Maple+1kConflict} & \#Solved & 269 & 218 & 174 & 53 & 220 & 934\\
& \#Sat + \#Unsat & 122+147 & 93+125 & 73+101 & 7+46 & 105+115 &\\
& Mean time & 1019s & 849s & 982s & 1659s & 890s &\\ 
&Reduction ratio& 22.15\% & 20.83\% & 25.50\% & 6.87\% & 17.92\% &\\ \hline
{\bf Maple+1.5kConflict} & \#Solved & 267 & 219 & 172 & 55 & 225 & 938\\
& \#Sat + \#Unsat & 122+145 & 95+124 & 72+100 & 8+47 & 109+116 &\\
& Mean time & 1030s & 766s & 983s & 1860s & 850s &\\ 
&Reduction ratio& 22.39\% & 20.64\% & 27.06\% & 6.37\% & 19.04\% &\\ \hline
{\bf Maple+eReduceDB} & \#Solved & 263 & 216 & 174 & 54 & 228 & 935\\
& \#Sat + \#Unsat & 118+145 & 94+122 & 72+102 & 8+46 & 110+118 &\\
& Mean time & 1061s & 842s & 990s & 1721s & 911s &\\ 
&Reduction ratio& 25.13\% & 23.47\% & 30.07\% & 8.56\% & 22.51\% &\\ \hline
{\bf Maple+eR} & \#Solved & 262 & 223 & 170 & 43 & 220 & 918\\
& \#Sat + \#Unsat & 117+145 & 98+125 & 70+100& 7+36 & 106+114 &\\
& Mean time & 984s & 821s & 907s & 1586s & 865s &\\ 
& Reduction ratio &18.49\% & 16.95\% & 21.73\% & 4.25\% & 14.52\%& \\ \hline
\end{tabular}
\caption{\small Comparison of different learnt clause vivification activation strategies on instances of recent SAT competitions. The cutoff time is 5000 seconds for each solver and instance.}
\label{restartStrategy}
\end{table}

\section{Improvements in Function $liveClause(C)$  \label{approachImprovement}}

The proposed clause vivification is implemented in a SAT solver by defining three functions: $liveRestart(nbNewLearnts, \sigma)$, $liveClause(C)$ and $sort(C)$. The results in Table~\ref{simplifyFrequentApp} and~\ref{restartStrategy} appear to indicate that the performance of the proposed vivification is not very sensitive to the different definitions of $liveRestart(nbNewLearnts, \sigma)$, and that keeping the current literal order in a clause that has to be vivified is the best option. In this section, we focus on function $liveClause(C)$  in order to improve the clause vivification of Maple+. In particular, we are interested in analyzing if it is worth re-vivifying clauses and vivifying original clauses during pre- and inprocessing.
We consider Maple+ because it was ranked first in the main track of SAT Competition 2017\footnote{Maple+ is called Maple\_LCM in the SAT Competition, where LCM stands for Learnt Clause Minimization.}. Thus, our goal is to further improve the results of SAT Competition 2017.

\subsection{Vivifying clauses more than once}

A common feature of the previous definitions of $liveClause(C)$ is that the function returns false if clause $C$ was already vivified in a previous call of Algorithm~\ref{vivifying}. However, as the search proceeds, and especially as more learnt clauses are added to the clause database, more redundant literals can be detected in $C$ by unit propagation so that $C$ can be vivified again. Since clause vivification is time-consuming, we have to define relevant and precise conditions under which a solver can re-vivify $C$.

The LBD value of $C$ can be considered  to be an estimation of the number of decisions that are needed to falsify $C$. We use the decrease of the LBD value of $C$ to measure the probability that unit propagation detects more redundant literals in $C$: If the LBD value of $C$ is decreased, then fewer decisions can be needed to falsify $C$ and unit propagation can detect more redundant literals in $C$. Concretely,
in Maple+, as in Glucose, every time $C$ is used to derive a new learnt clause, a new LBD value of $C$ is 
computed. Let $d_0$ be the LBD value of $C$ at time $t_0$. We say that the LBD value of $C$ is decreased 
$\lambda$ times since $t_0$ if $C$ is used to derive a new learnt clause at times $t_1, t_2, \ldots, t_\lambda$ 
($t_0<t_1< \cdots< t_\lambda$) and $d_0 > d_1 > \cdots > d_\lambda$, where  $d_i$ ($1\leq i \leq \lambda$) 
is the LBD value of $C$ computed at time $t_i$.

Note that the actual computed LBD value depends on the quality and ordering of the decisions, which are continuously improved as the search proceeds. So, when the computed LBD value is decreased one time, it may be only due to an improvement of the quality and/or ordering of the decisions. So,
before allowing to vivify $C$ again, we require that the LBD of $C$ is decreased 2 times since the last time $C$ was vivified. Function liveClause+$(C)$ described in Algorithm~\ref{newLiveClause} implements this strategy.

\begin{algorithm}[!htb]
\KwIn{A clause $C$}
\KwOut{$true$ or $false$}
\Begin{
	\If {$C$ is a learnt clause in CORE or TIER2} {
		\If {$C$ was never before vivified or the LBD of $C$ is decreased 2 times since its last vivification by Algorithm~\ref{vivifying}} {
			return $true$;
		}
	}
	return $false$;
}
\caption{liveClause+$(C)$: select a clause $C$ to vivify \label{newLiveClause}}
\end{algorithm}

The incorporation of function liveClause+$(C)$ into Maple+ results in a new solver called Maple++.

\subsection{Vivifying original clauses}


Another common feature of the previous definitions of $liveClause(C)$ is that the function returns true only if $C$ is a learnt clause. However, the original clauses of a SAT instance can also contain redundant literals. The question is whether to vivify all original clauses during preprocessing, and whether to vivify or re-vivify all original clauses during inprocessing.

Vivifying each original clause, even limited to preprocessing, is a time-consuming task in large instances. This task could be notably speeded up by using trie data structures as in~\cite{HS07}. We conducted preliminary experiments to compare two preprocessing vivifications (see Section \ref{SectAnaCVall}): The first one stops the vivification when a fixed number of propagated literals is reached, and the second one vivifies all original clauses without any limit. Without considering the time for preprocessing and with a cutoff time of 5000 seconds for the search, the solver with the two preprocessing vivifications solves almost the same number of instances, meaning that it is not worth vivifying every original clause when an instance is very large.


So, we empirically limit the number of propagated literals in preprocessing vivification to $10^8$. To present the selection of original clauses that will be vivified during the search process, we define the notions of {\em useful conflict} and {\em useful clause}. A conflict during the search process is said to be {\em useful} if the LBD of the learnt clause derived from that conflict is smaller than or equal to $\gamma$, where $\gamma$ is an integer parameter that is fixed empirically to 20 (see Section \ref{SectRobust20}). Intuitively, if the LBD is greater than 20, the conflict probably needs more than 20 decisions to be produced, so that its reproduction is improbable during the search process. We believe that only useful conflicts contribute to solving an instance. Thus, the solver only needs to focus on useful conflicts. Consequently, we say that a clause is {\em useful} if it was used to derive a learnt clause in a useful conflict.

Our plan is to vivify every useful original clause and re-vivify it under conditions similar to those applied to learnt clauses. Algorithm~\ref{finalLiveClause} defines the function LiveClause++$(C)$ that we will use to improve Maple+.

\begin{algorithm}[!htb]
\KwIn{A clause $C$}
\KwOut{$true$ or $false$}
\Begin{
	\If {$C$ is a learnt clause} {
		\If {$C$ is in CORE or TIER2} {
			\If {$C$ was never  before vivified, or the LBD of $C$ is decreased 2 times since its last vivification by Algorithm~\ref{vivifying}, or the LBD of $C$ is decreased to 1 since its last vivification by Algorithm~\ref{vivifying}} {
				return $true$;
			}
		}
		return $false$;
	}
	\Else {
		\If {$C$ was used to derive at least one learnt clause of LBD smaller than or equal to 20 since the last execution of Algorithm~\ref{vivifying}} {
			\If {$C$ was never  before vivified, or the LBD of $C$ is decreased 3 times since its last vivification by Algorithm~\ref{vivifying}, or the LBD of $C$ is decreased to 1 since its last vivification by Algorithm~\ref{vivifying}} {
				return $true$;
			}
		}
		return $false$;
	}
}
\caption{liveClause++$(C)$: select a clause $C$ to vivify \label{finalLiveClause}}
\end{algorithm}

The LBD of an original clause is initialized to the number of literals it contains. Then, as in learnt clauses, every time the original clause is used to derive a new learnt clause, a new LBD value is computed. Since an original clause presumably contains fewer redundant literals than a learnt clause, liveClause++$(C)$ requires that the LBD of an original clause is decreased one more time (3) than a learnt clause since its last vivification to vivify it again. 

A particular case is when the LBD of a learnt or original clause becomes 1 since its last vivification. Obviously, the LBD value cannot  be decreased below 1. Moreover, a clause with LBD 1 is likely very powerful in unit propagation, because all its literals are asserted within one decision level. Then, when the LBD of a clause is decreased to 1 since its last vivification, the clause will be re-vivified regardless of how many times the LBD value is decreased.

Since both learnt and original clauses are vivified, it is necessary to establish the order in which such clauses are considered. In our approach, Algorithm~\ref{vivifying} vivifies learnt clauses before original clauses, because learnt clauses presumably contain more redundant literals.

The incorporation of function liveClause++$(C)$ in Maple+ results in a new solver called Maple\_CV, standing for MapleSAT with clause vivification. Thus, Maple\_CV is Maple+ but uses liveClause++$(C)$ to select learnt and original clauses to vivify or re-vivify.

We add a preprocessing in Maple\_CV that vivifies every original clause using Algorithm~\ref{vivifying} but vivification is stopped when the number of propagated literals reaches $10^8$. We then obtain our final solver called  Maple\_CV+ in this paper.

\subsection{Empirical evaluation}

In this subsection, we first show the performance of our final solver Maple CV+, providing empirical evidence for the effectiveness of our approach. Then, we analyze the behaviour of Maple\_CV+. Finally, we show the robustness of our approach by comparing Maple\_CV+ with several variants and Cadical (version 2018)~\cite{Biere2018}. Note that Cadical is a non MiniSat-based solver and the used version implements learnt clause vivification in the spirit of the ideas presented in~\cite{LLXML17}.

As in Table \ref{restartStrategy}, the test suite includes the 1450 instances from  the main track (application + crafted) of the SAT Competition 2014, 2016 and 2017. The experiments were performed  on Intel Xeon E5-2680 v4 processors at 2.40GHz and 20GB of memory under Linux. The cutoff time is 5000 seconds for each solver and instance, including the preprocessing time and the search time, unless otherwise stated.

\subsubsection{Effectiveness of the proposed approach}
We compare Maple+, Maple++ and Maple\_CV to evaluate the improvements of Algorithm~\ref{newLiveClause} (Function liveClause+$(C)$) and Algorithm~\ref{finalLiveClause} (Function liveClause++$(C)$). Maple++ and Maple\_CV are like Maple+ but use a different definition of liveClause$(C)$ to select the clauses that will be vivified by Algorithm~\ref{vivifying}. The final solver Maple\_CV+ is also included in the comparison.

\begin{table}[htp]
\centering
\begin{tabular}{|ll|l|l|l|l|l|l|}\hline
&  & \multicolumn{2}{c}{SAT 2014} & \multicolumn{2}{|c|}{SAT 2016} & SAT 2017 & Total\\
\cline{3-7}
 &                                                                       & application  & crafted      & application & crafted   & main          &\\  
 Solver                               &                              & 300 ins.      & 300 ins.     & 300 ins.      & 200 ins  & 350 ins.     & 1450 ins.\\\hline
{\bf Maple+}                      & \#solved                 & 265            & 218 & 175 & 56               & 219        & 933\\
                                         & \#Sat + \#Unsat     & 121+144    & 93+125      & 74+101       & 8+48     &103+116      &\\
                                         & Mean Time             & 1054s        & 779s           &1023s          &1619s     & 878s          &\\ \hline
{\bf Maple++}                   & \#solved                   & 264           & 221              & 174           & \bf 60       & 225            & 944\\
                                        & \#Sat + \#Unsat        & 118+146   & 96+125        & 72+102     & 9+51       & 108+117    &\\
                                         & Mean Time              & 1018s       & 790s            & 857s          & 1695s     & 942s           &\\ \hline                                         
{\bf Maple\_CV}          & \#solved                  & \bf 271       & 221               &  176           &  55          &  232          & 955 \\
                                         & \#Sat + \#Unsat      & 124+147    &  96+125        & 73+103      & 6+49       & 110+122   &\\
                                        & Mean Time              &982s           & 783s              & 1031s        & 1680s      & 1052s       &\\ \hline                                                                        
{\bf Maple\_CV+}       & \#solved                 & 269              & \bf 222                 & \bf 179     & 58              & \bf 239     & \bf 967\\
                                       & \#Sat + \#Unsat      & 120+149      & 97+125          & 74+105    & 7+51          & 115+124  &\\
                                       & Mean Time             & 1012s           & 804s              & 1037s      & 1671s        & 1032s      &\\\hline
\end{tabular}
\caption{\small Effectiveness of re-vivifying clauses and vivifying original clauses\label{improvements}}
\end{table}

Table~\ref{improvements} shows the results of Maple+, Maple++, Maple\_CV and  Maple\_CV+. Several observations can be made from these results:

\begin{itemize}
\item Re-vivifying a learnt clause when its LBD value is decreased two times as done in Maple++ is effective. In fact, Maple++ solves 11 instances more than Maple+.

\item Vivifying original clauses during the search, using function liveClause++$(C)$ to  select them, is effective in terms of the total number of instances solved within the cutoff time. In fact, Maple\_CV solves 11 instances more than Maple++ and 22 instances more than Maple+.

\item The incorporation of a limited preprocessing to vivify original clauses by unit propagation further improves the performance of the solvers: Maple\_CV+ solves 23 instances more than Maple++ and 34 instances more than Maple+.
\end{itemize}

These results clearly indicate that vivification of both learnt and original clauses during pre- and inprocessing is a decisive solving technique for boosting the performance of modern CDCL SAT solvers.

It is important to highlight that Maple+ was ranked first in the main track of SAT Competition 2017, and that Maple\_CV+ solves 8, 6 and 20  instances more than Maple+ in the main track of SAT Competition 2014, 2016 and 2017, respectively. To assess the significance of this result, we would like to mention that the number of additional solved instances in the main track of consecutive SAT competitions is typically less than 5 instances. 

\subsubsection{Analysis of the behaviour of Maple\_CV+}
\label{SectAnaCVall}

Recall that Maple\_CV+ vivifies each original clause during preprocessing but stops preprocessing vivification when the number of propagated literals reaches $10^8$. Table \ref{preprocessing} compares Maple\_CV+ with its variant Maple\_CV+all, which is Maple\_CV+ but never stops the preprocessing vivification. In the table, the preprocessing time of the two solvers is not limited but the search time (after the preprocessing) is limited to 5000 seconds.

\begin{table}[htp]
	\centering
	\begin{tabular}{|ll|l|l|l|l|l|l|}\hline
		&  & \multicolumn{2}{c}{SAT 2014} & \multicolumn{2}{|c|}{SAT 2016} & SAT 2017 & Total\\
		\cline{3-7}
		&  & application  & crafted & application  & crafted & main &\\ 
		Solver          &   & 300 ins.      & 300 ins. & 300 ins.     & 200 ins. & 350 ins. & 1450 ins.\\ \hline
		
		{\bf Maple\_CV+}       & \#solved                 & 269              & 222                 &  179     & 58              & 239     &  967\\
                                       & \#Sat + \#Unsat      & 120+149      & 97+125          & 74+105    & 7+51          & 115+124  &\\

		& preprocessing time & 3.9s & 2.1s &7.5s &63.9s  & 6.5s &\\ 
		& search time & 994.4s & 826.3s &1237.7s &1619.7s  & 1001s & \\
		\hline
		
		{\bf Maple\_CV+all} & \#Solved & 270 & 219 & 180 & 56 & 239 & 964\\
		& \#Sat + \#Unsat & 123+147 & 95+124 & 71+109& 5+51 & 113+126 &\\
		& preprocessing time& 494.6s & 34.3s & 203.4s & 62.86s & 64.9s &\\ 
		& search time & 1009.8s & 848.5s &1186s &1635.8s  & 1032s & \\
		\hline
		
	\end{tabular}
	\caption{\small Results of Maple\_CV+ and Maple\_CV+all on  instances of recent SAT competitions}
	\label{preprocessing}
\end{table}

Maple\_CV+ solves almost the same number of instances as Maple\_CV+all with similar search time and substantially shorter preprocessing time. Note that the two solvers vivify the original clauses one by one in their natural order during preprocessing. When vivifying all original clauses during preprocessing is too time-consuming, it might be more profitable to devise a clever heuristic to select and order a subset of original clauses to vivify, than to invest in implementing a more efficient vivification algorithm.

Table~\ref{reductionRateOrigineLearnt} compares the reduction ratios of original clauses and learnt clauses in the solver Maple\_CV+,  measured as $(a-b)/a\times 100$, where $a$  is the total number of literals in all the original (learnt) clauses $C$ such that liveClause++$(C)$ is true before applying clause vivification and $b$ is the total number of literals in the vivified original (learnt) clauses. The displayed data are averaged among the solved instances in each group. We observe that the reduction ratio of original clauses is substantially smaller than the reduction ratio of learnt clauses, because original clauses presumably contain fewer redundant literals. However, even with these small reduction ratios, the results of Table~\ref{improvements} indicate that vivifying the original clauses selected by the function liveClause++$(C)$ is effective, except for the very special crafted instances of SAT Competition 2016 for which the reduction ratio of original clauses is too low. 

\begin{table}[htp]
\centering
\begin{tabular}{|c|c|c|c|c|c|}\hline
Reduction ratio     & \multicolumn{2}{c}{SAT 2014} & \multicolumn{2}{|c|}{SAT 2016} & SAT 2017 \\
\cline{2-6}
                             & application    & crafted           & application & crafted                & main \\
                             & 300 ins.         & 300 ins.          & 300 ins.     &  200 ins.              & 350 ins.\\ \hline
Original clauses   & 2.71\%          & 2.85\%            & 3.98\%      & 1.11\%                & 1.74\%\\ \hline
Learnt clauses     & 22.35\%        & 19.67\%          & 25.69\%     & 9.05\%               & 18.85\% \\ \hline
\end{tabular}
\caption{\small Reduction ratios of original and learnt clauses that were vivified in Maple\_CV+ on instances of recent SAT competitions.\label{reductionRateOrigineLearnt}}
\end{table}

When a clause is involved to derive a learnt clause from a conflict, its LBD is re-computed. Table \ref{reductionLBDoriginalLearnt} shows the percentage of original (learnt) clauses whose LBD is decreased among the original (learnt) clauses whose LBD is re-computed, as well as the percentage of these clauses with decreased LBD that are re-vivified in Maple\_CV+. The displayed data are averaged among the solved instances in each group. The percentage of the re-vivified original clauses is high, partly because the LBD value of many original clauses is reduced to 1 after their last vivification. The results show that once the LBD of a clause is decreased, its probability to be re-vivified is high, suggesting that the reduction of LBD values could be a relevant indicator of the search progression that deserves more investigation in the future. 

\begin{table}[htp]
\centering
\begin{tabular}{|c|c|c|c|c|c|}\hline
       & \multicolumn{2}{c}{SAT 2014} & \multicolumn{2}{|c|}{SAT 2016} & SAT 2017 \\
       \cline{2-6}
                             & application    & crafted           & application & crafted                & main \\
                             & 300 ins.         & 300 ins.          & 300 ins.     &  200 ins.              & 350 ins.\\ \hline
\% original clauses with reduced LBD  & 2.7\%   & 4.7\%    & 3.8\%    & 0.79\%   & 3.3\%\\ \hline
\% of re-vivified original clauses with reduced LBD  & 35.4\%   & 47.8\%    & 53.4\%    & 50.4\%  & 47.6.\%\\ \hline
\% learnt clauses with reduced LBD     & 4.4\%        & 4.6\%   & 4.4\%   & 8.8\%   & 3.9\% \\ \hline
\% of re-vivified learnt clauses with reduced LBD  & 25.6\%   & 23.4\%    & 35.1\%  & 18.4\%  & 25.3.\%\\ \hline
\end{tabular}
\caption{\small Percentage of original and learnt clauses whose LBD values were decreased in Maple\_CV+ on instances of recent SAT competitions, and percentage of re-vivified original and learnt clauses with decreased LBD.\label{reductionLBDoriginalLearnt}}
\end{table}

From Table \ref{cases}, Table \ref{restartStrategy}, Table \ref{reductionRateOrigineLearnt} and Table \ref{reductionLBDoriginalLearnt}, we can get an explanation why our approach is not so effective for the 200 crafted instances in SAT competition 2016: these instances have the highest percentage (51.3\%) of clauses that are checked by clause vivification but not simplified, i.e., no redundant literal is detected in these clauses (Table \ref{cases}), and the lowest reduction ratio (Table \ref{restartStrategy} and Table \ref{reductionRateOrigineLearnt}). More importantly, only 0.79\% of original clauses in these instances have their LBD value reduced during the search (Table \ref{reductionLBDoriginalLearnt}). On the contrary, these tables also explain why our approach is effective for all other groups of instances.

\subsubsection{Robustness of the clause vivification approach \label{SectRobust20}}

\begin{figure}
    \centering
\begin{minipage}{0.6\linewidth}
    \includegraphics[width=\textwidth]{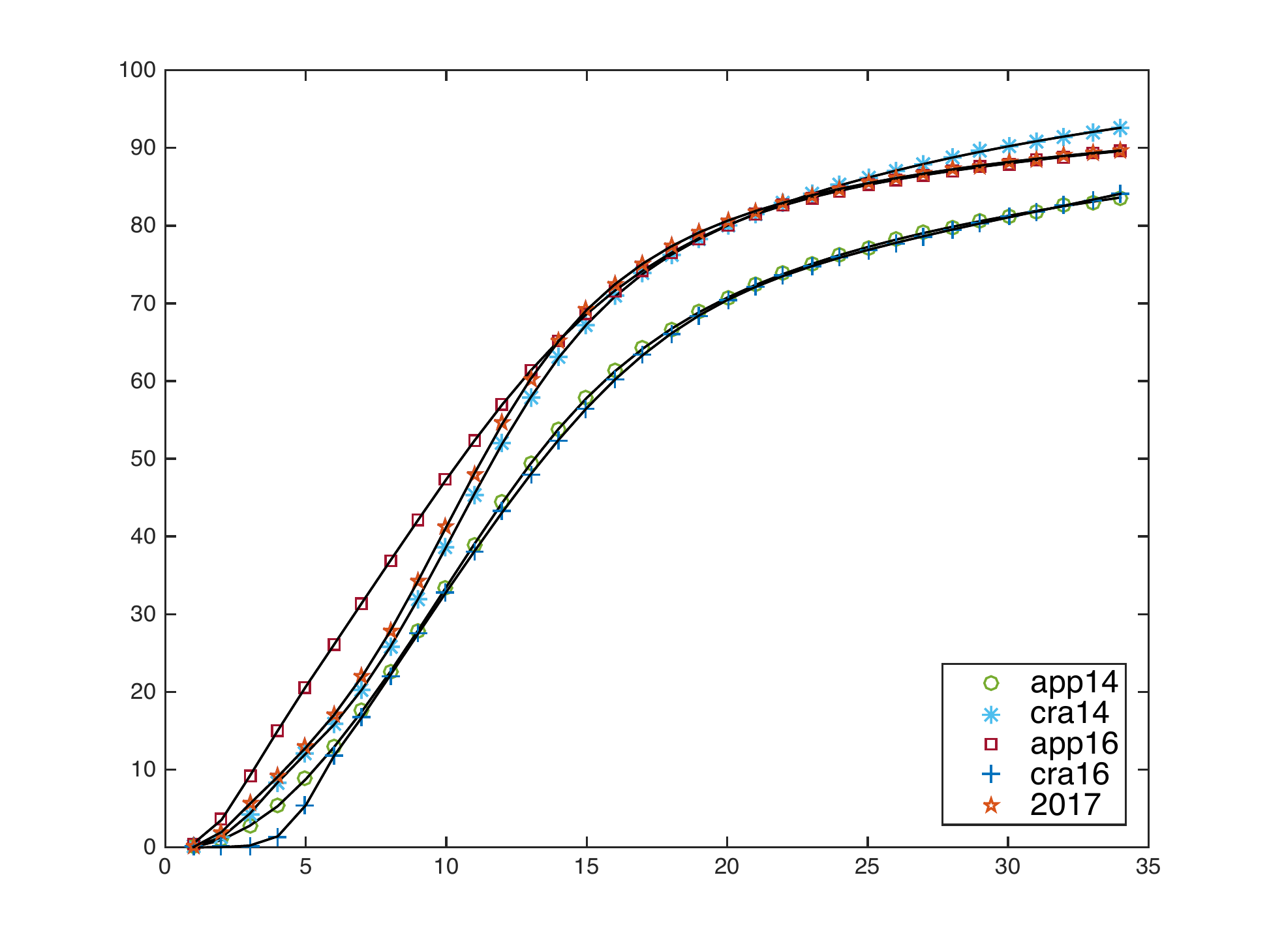}
\end{minipage}
\caption{\small Cumulative distribution functions for LBD of learnt clauses generated by Maple+ on benchmarks of SAT competition 2014, 2016 and 2017. Each point $(x, y)$ represents the percentage $y\%$ of learnt clauses with LBD smaller than or equal to $x$, averaged among all the solved instances in a group.}
 \label{lbd_cum_distribution}
\end{figure}

Recall that Maple\_CV+ vivifies each original clause used to derive a learnt clause with LBD smaller than or equal to $20$. This constant $20$ comes from the study of the cumulative distribution of the LBD of learnt clauses generated by Maple+ during the search process, showed in Figure~\ref{lbd_cum_distribution}. Each point $(x, y)$ of the figure shows the percentage $y\%$ of learnt clauses with the LBD value smaller than or equal to $x$, averaged among the solved instances in each group from the main (industrial + hard combinatorial) track of SAT Competition 2014, 2016 and 2017. Our purpose is to devise a global constant effective enough for all the instances, so that one does not need to adjust the value for any special family of instances. 

It is interesting to note that the cumulative distribution function has a similar form in all the groups of instances, and the LBD values around $20$ roughly corresponds to a transition zone such that, for each LBD value smaller than the values of this zone, there are many learnt clauses and, for each LBD value greater than the values of this zone, there are fewer learnt clauses, meaning that a conflict with LBD greater than $20$ has little chance to be reproduced. 

Table \ref{robustness} compares the value $20$ with $15$ and $25$. In the table Maple\_CV+ with value $\lambda\in \{15, 20, 25\}$ means that Maple\_CV+ vivifies each original clause used to derived a learnt clause with LBD smaller than or equal to $\lambda$. The results confirm that it is relevant to distinguish useful and non-useful conflicts and $20$ is a good value for that, because Maple\_CV+
with value $20$ is significantly better than  Maple\_CV+ with value $15$ or $25$, and the three variants of Maple\_CV+ perform better than the base solvers Maple+ and Maple++ from which Maple\_CV+ is developped, showing the robustness of our approach.

\begin{table}[htp]
\centering
\begin{tabular}{|ll|l|l|l|l|l|l|}\hline
&  & \multicolumn{2}{c}{SAT 2014} & \multicolumn{2}{|c|}{SAT 2016} & SAT 2017 & Total\\
\cline{3-7}
 &                                                                       & application  & crafted      & application & crafted   & main          &\\  
 Solver                               &                              & 300 ins.      & 300 ins.     & 300 ins.      & 200 ins  & 350 ins.     & 1450 ins.\\\hline    
 {\bf Maple+}                      & \#solved                 & 265            & 218 & 175 & 56               & 219        & 933\\
                                         & \#Sat + \#Unsat     & 121+144    & 93+125      & 74+101       & 8+48     &103+116      &\\
                                         & Mean Time             & 1054s        & 779s           &1023s          &1619s     & 878s          &\\ \hline
{\bf Maple++}                   & \#solved                   & 264           & 221              & 174           & \bf 60       & 225            & 944\\
                                        & \#Sat + \#Unsat        & 118+146   & 96+125        & 72+102     & 9+51       & 108+117    &\\
                                         & Mean Time              & 1018s       & 790s            & 857s          & 1695s     & 942s           &\\ \hline   
                                                               
{\bf Maple\_CV+}        & \#solved                  &  264           & 218               &  178           &  56          &  231          & 947 \\

 with value 15            & \#Sat + \#Unsat      & 116+148    &  93+125        & 71+107      & 7+49       & 109+122   &\\
                                        & Mean Time              &947s           & 767s              & 986s        & 1543s      & 933s       &\\ \hline                                         
                                        
{\bf Maple\_CV+}       & \#solved                 & \bf 269              & 222                 & \bf 179     & 58              & \bf 239     & \bf 967\\
 with value 20                & \#Sat + \#Unsat      & 120+149      & 97+125          & 74+105    & 7+51          & 115+124  &\\
                                       & Mean Time             & 1012s           & 804s              & 1037s      & 1671s        & 1032s      &\\ \hline                                        
                                                          
{\bf Maple\_CV+}       & \#solved                 & 266              & \bf 223           & 177         & 58              & 233          & 957\\
 with value 25                                      & \#Sat + \#Unsat      & 120+146      & 97+126          & 73+104    & 8+50          & 111+122  &\\
                                       & Mean Time             & 1025s           & 832s              & 1003s      & 1662s        & 968s       &\\ \hline

\end{tabular}
\caption{\small Comparison of three variants of Maple\_CV+ with the base solvers Maple+ and Maple++  on instances of recent SAT competitions.\label{robustness}}
\end{table}

All the solvers considered so far have been derived from MiniSat by incorporating a number of improvements that allow them to outperform MiniSat. To show that the proposed vivification is also suitable for other non MiniSat-based SAT solvers, we conducted an experiment with three different variants of the SAT solver Cadical-2018~\cite{Biere2018}, which was created by Armin Biere and incorporates inprocessing techniques such as probing, subsumption and bounded variable elimination. Cadical-2018 has been submitted to SAT Competition 2018. A previous version of Cadical
(Cadical-2017~\cite{Biere2017}) performed already better than the solver Lingeling by the same author in which clause vivification is missing. In fact, in SAT competition 2017, Cadical-2017 solved 31 instances more than Lingeling among the 350 instances of the main track. Cadical-2017 included inprocessing vivification restricted to irredundant clauses. Cadical-2018 includes, in addition, inprocessing vivification applied to redundant clauses implemented in the spirit of the ideas presented in our IJCAI paper \cite{LLXML17}. 
Here, redundant clauses roughly correspond to the learnt clauses that do not subsume any original clause.

We compared Cadical-2018 with the following variants:

\begin{itemize}
\item Cadical-2017.
\item Cadical-2018 without any clause vivification (i.e., clause vivification is disabled).
\item Cadical-2018 with clause vivification only  applied to irredundant clauses.
\item Cadical-2018 with clause vivification only  applied to redundant clauses.
\end{itemize}

Table~\ref{cadical} shows the experimental results. We observe that Cadical-2018 solves 23 instances more than the variant without any clause vivification, 19 instances more than the variant only applying clause vivification to irredundant clauses and 11 instances more than the variant only applying clause vivification to redundant clauses. These results provide another evidence that it is important to vivify both original and learnt clauses, as well as that the proposed vivification is also well suited for non MiniSat-based SAT solvers.

 
\begin{table}[htp]
\centering
\begin{tabular}{|ll|l|l|l|l|l|l|}\hline
&  & \multicolumn{2}{c}{SAT 2014} & \multicolumn{2}{|c|}{SAT 2016} & SAT 2017 & Total\\
\cline{3-7}
 &                                                                       & application  & crafted      & application & crafted   & main          &\\  
 Solver                               &                              & 300 ins.      & 300 ins.     & 300 ins.      & 200 ins  & 350 ins.     & 1450 ins.\\\hline
{\bf Cadical-2017}      & \#solved                 & 240            & 179 & 169 & 60  & 197  & 845\\
                              & \#Sat + \#Unsat     & 103+137    & 86+93  & 68+101       & 5+55   &87+110      &\\
                                       \hline
                                        
{\bf Cadical-2018}   & \#solved     &  243   &  196   &  170     &  72     &  229    &  910 \\
without any vivification & \#Sat+\#Unsat      & 104+139    & 90+106        & 74+96       & 9+63    & 114+115     &\\
									\hline                                        
                                        
{\bf Cadical-2018} with         & \#solved         & 246           & 199      & 171     &  71      & 227          & 914\\
 vivification applied to            & \#Sat+\#Unsat  & 104+142   & 90+109 & 74+97 & 7+64 & 115+112    &\\
irredundant clauses only        &                        &                     &            &            &           &                      &\\         
 \hline
                                         
{\bf Cadical-2018} with                   & \#solved     & 246   & 202   & 170     &  72     & 232    &  922 \\
 vivification applied to	& \#Sat+\#Unsat      & 102+144         & 94+108 & 74+96       & 8+64    & 115+117     &\\
 redundant clauses only         &                               &           &            &            &           &                      &  \\               
									\hline                                                                                      
{\bf Cadical-2018}      & \#solved      & \bf 251     & \bf 204      & \bf 173    & \bf 73      & \bf 232        & \bf 933 \\
                                         & \#Sat+\#Unsat      & 105+146    & 92+112        & 74+99       & 10+63    & 115+117     &\\
                                        \hline

\end{tabular}
\caption{\small Comparison of Cadical-2018 with its three variants without clause vivification or with weaker clause vivification
  on instances of recent SAT competitions.\label{cadical}}
\end{table}

\section{Conclusions} \label{conclusions}
 We defined an original inprocessing clause vivification approach that eliminates redundant literals of original and learnt clauses by applying unit propagation, and that can also be applied during preprocessing to vivify original clauses. We also performed an in-depth empirical analysis that shows that the proposed clause vivification is robust and allows to solve a remarkable number of additional instances from recent SAT competitions. The first part of the experimentation is devoted to show that learnt clause vivification by itself is a powerful solving technique that boosts the performance of the best state-of-the-art SAT solvers. The second part of the experimentation is devoted to show that the combination of learnt clause vivification and original clause vivification leads to a yet more powerful 
 solving technique.
 
The clause vivification approach proposed in this paper allowed us to win the gold medal of the main track in SAT competition 2017 and the bronze medal of the main track in SAT competition 2018. Furthermore, the best 13 solvers of the main track of SAT Competition 2018,  developed by different author sets,
all use our learnt clause vivification approach.\\
 

{\bf \noindent Acknowledgements}
\\
We thank Armin Biere for explaining us how clause vivification is implemented in Cadical-2018. This work was partially supported by the National Natural Science Foundation of China (Grants no. 61370183, 61370184, 61472147), the Matrics Platform of the Universit\'e de Picardie Jules Verne, the project LOGISTAR from the EU H2020 Research and Innovation Programme under Grant Agreement No.\ 769142 and the MINECO-FEDER project RASO TIN2015-71799-C2-1-P.
\\


\bibliographystyle{elsarticle-num}
\bibliography{artint2New_2018.7.20}
\end{document}